\documentclass[10pt,lettersize,journal]{IEEEtran}
\usepackage{amsmath,amsfonts}
\usepackage{algorithmic}
\usepackage{algorithm}
\usepackage{array}
\usepackage[caption=false,font=normalsize,labelfont=sf,textfont=sf]{subfig}
\usepackage{textcomp}
\usepackage{stfloats}
\usepackage{url}
\usepackage{verbatim}
\usepackage{graphicx}
\usepackage[utf8]{inputenc}
\usepackage{amssymb}
\usepackage{amsmath}
\usepackage{adjustbox}
\usepackage{multirow}
\usepackage{bm}
\usepackage{xcolor}
\usepackage{siunitx}
\usepackage{setspace}
\usepackage{colortbl}
\usepackage{color, colortbl}
\definecolor{Orange}{rgb}{0.0,0.8,0.1}
\definecolor{red}{rgb}{1.0, 0.0, 0.0}
\definecolor{LightOrange}{rgb}{0.8,0.7,0.1}
\definecolor{green}{rgb}{0.0, 1.0, 0.0}

\newcommand{\sao}[1]{{\color{black}#1}} 
\usepackage{cite}
\def\BibTeX{{\rm B\kern-.05em{\sc i\kern-.025em b}\kern-.08em
    T\kern-.1667em\lower.7ex\hbox{E}\kern-.125emX}}
\usepackage{balance}

\begin{document}

\title{\sao{SLYKLatent: A Learning Framework for Gaze Estimation Using Deep Facial Feature Learning}}

\author{Samuel~Adebayo,~\IEEEmembership{Member,~IEEE,}
        ~Joost~C.~Dessing, and 
        Se\'{a}n~McLoone~\IEEEmembership{Senior,~IEEE,}
\IEEEcompsocitemizethanks{\IEEEcompsocthanksitem Code Repository will be at https://github.com/exponentialR/SLYKLatent. The authors are with the Centre for Intelligent Autonomous Manufacturing Systems, Queen's University Belfast, Northern Ireland, United Kingdom.\protect\\
E-mail: \{sadebayo01, j.dessing, s.mcloone\}@qub.ac.uk
}

\thanks{********}}
\markboth{Journal of \LaTeX\ Class Files,~Vol.~14, No.~8, August~****}%
{Shell \MakeLowercase{\textit{et al.}}: A Sample Article Using IEEEtran.cls for IEEE Journals}


\maketitle

\begin{abstract}
In this research, we present SLYKLatent, a novel approach for enhancing gaze estimation by addressing appearance instability challenges in datasets due to aleatoric uncertainties, covariant shifts, and test domain generalization. SLYKLatent utilizes self-supervised learning for initial training with facial expression datasets, followed by refinement with a patch-based tri-branch network and an inverse explained variance weighted training loss function. Our evaluation on benchmark datasets achieves a $10.98\%$ improvement on Gaze360, supersedes the top result with $3.83\%$ improvement on MPIIFaceGaze, and leads on a subset of ETH-XGaze by $11.59\%$, surpassing existing methods by significant margins. Additionally, adaptability tests on RAF-DB and Affectnet show 86.4\% and 60.9\% accuracies, respectively. Ablation studies confirm the effectiveness of SLYKLatent's novel components. This approach has strong potential in human-robot interaction.
\end{abstract}

\begin{IEEEkeywords}
Human-Machine Interaction, Facial feature extraction, Gaze estimation, Facial Expression Recognition, Human-Robot Interaction, Intention inference, Computational Intelligence.
\end{IEEEkeywords}

\section{Introduction}\label{sec:introduction}
\subsection{Background and Motivation}
\IEEEPARstart{H}{uman-Robot} interaction (HRI) is an interdisciplinary field that focuses on the understanding, modelling, and design of interactions between humans and robots. HRI plays a critical role in many areas such as education, healthcare, manufacturing, and entertainment.  Irrespective of the form of interaction, the goal of any HRI framework is to create natural and effective interactions between humans and robots, where the robot can understand and respond to the human's needs, intentions, and goals.\cite{c1, c2, c3}. 

One crucial aspect of HRI is the ability to infer human intention, that is the capability of a robot to predict and deduce a human's future actions before or during task execution \cite{c4, c52, c53}. This is a vital component of HRI frameworks, as it enables the robot to compute and anticipate human intentions at every stage of interaction, leading to more efficient, safe, and effective interactions \cite{c5}. Therefore, there is a need to identify efficient methods to capture, analyze, and process the observable cues required to infer human intention. Additionally, it requires the development of advanced frameworks and techniques that can optimally represent human perspectives, activities, and intentions \cite{c5}. 

Within the HRI field, the predominant approach has been to concentrate on deciphering and encoding human verbal cues to determine an appropriate response from a robot. However, in this framework, non-verbal cues from humans are frequently neglected and not given sufficient consideration. Understandably, since HRI is a replication of human-to-human interaction, it is reasonable to want to emulate a large part of the human communication process for a HRI framework \cite{c6}. Human-to-human interactions rely not only on verbal cues, but also on posture, gesture cues, and facial cues, all of which may carry information pertaining to the executed task or action (i.e. intention) \cite{c6, c7}. Humans can achieve this by understanding and interpreting observable patterns in the available information from the aforementioned cues. Here we will focus on a subset of these cues: facial cues.

Recent developments in HRI systems proposed the fusion of two or more visual cues, such as hand, eye, and interacting object, for inferring human intention \cite{c4, c54, c55} Other methods also suggest facial cues as a way of encoding expressiveness \cite{c8, c9, c10, c11}. While the combination of these visual signals to classify intention is not within the scope of this research, it is likely that the fusion of facial features such as gazes and facial expressions with other obtainable visual cues would improve human intention inference in an HRI framework \cite{c12, c6}.

Facial cues can be considered perceptible physiological characteristics located on the human face that can be analysed and converted into facial expressions, gestures, and gaze estimation, thereby facilitating additional forms of non-verbal communication. The computation tasks involved in vision-based gaze and facial expression estimation exhibit areas of overlap, as both entail the examination of facial images, either in their entirety or in part. Consequently, advancements in facial feature estimation have the potential to yield enhancements in gaze estimation capabilities.

Gaze estimation refers to the process of determining the direction of a person's gaze - it is a distinct aspect of facial feature estimation, and has received considerable attention in HRI research \cite{c13, c14, c30, c31, c32}.
Gaze direction is considered a crucial cue for human attention and intention, because of the tight link between attention, intention, and gaze control \cite{c15, c16, c69}. In addition, gaze direction is more universally understood than verbal cues, making it applicable across different cultures regardless of language. However, despite recent progress in utilizing learning-based techniques for extracting facial features, the practical implementation of gaze estimation often faces limitations due to the model's inability to accurately infer useful facial feature points from visual appearance in real-life scenarios. This limitation primarily arises from the challenges posed by appearance uncertainties and the need for domain generalization.
Appearance uncertainty pertains to the variations in lighting conditions, facial pose, expressions, and other factors that can influence the visual appearance of the face and subsequently impact the estimation of gaze direction. On the other hand, domain generalization refers to the model's capacity to perform effectively on new, unseen data from diverse domains, distinct from the training data it was exposed to.

The task of gaze estimation in the field of human-robot interaction is challenging due to various sources of uncertainty present in real-world scenarios. One of the major sources of uncertainty is heteroscedastic noise, which refers to variations in the level of noise across different samples. In the context of gaze estimation, this type of noise can arise from variations in lighting conditions, facial expressions, and other factors that affect the appearance of the face and gaze direction \cite{c18, c19}. This leads to increased errors in the prediction of gaze direction.

Another significant challenge in gaze estimation is equivariance shift, which refers to variations in the position, orientation, and scale of the face in the image. These changes can cause the associated labels to vary, resulting in a domain adaptation problem when dealing with unseen data and unaccounted variations in the feature space. \sao{Domain adaptation can be defined as the process by which a model that has been trained in one domain, or data distribution, is adapted to perform well in another typically distinct domain. \sao{This is important in gaze estimation as real-world conditions can vary widely, for instance in the presence of uncertainties in appearance or other forms of noise. Therefore, models must perform consistently across different contexts where such stochastic shifts can arise.}} Most deep neural networks exhibit invariance but lack equivariant capabilities, indicating that they can correctly predict or estimate the gaze point regardless of the face's position in the image. However, they are sensitive to changes in face orientation \cite{c20}. Consequently, the learned visual representations have a lower ratio of equivalence-to-shift unless the network architecture inherently computes equivariance shifts \cite{c20}.

To address these challenges, a robust and efficient approach to gaze estimation is needed to handle domain adaptation and appearance uncertainties. In this paper, we propose the Self Learn Your Key Latent (SLYKLatent) features framework, a combination of self-supervised learning and transfer feature fine-tuning. \sao{Self supervised learning (SSL) is a technique where the model is trained using automatically generated labels derived from the input data itself, allowing it to learn representations without the need for manually labelled data. Transfer feature fine-tuning, on the other hand, involves adapting models which have been pre-trained on large, diverse datasets to a specific target task, in this case, gaze estimation. This accelerates the training process and is particularly effective when there the target task training data is limited in size and diversity.} By using a crowdsourced dataset that includes a diverse range of participants from different backgrounds, facial expressions, and non-facial images, our framework is able to adapt to different lighting conditions and appearance variations. Additionally, by introducing equivariance through the use of appearance transformations in self-supervised pretraining, our framework extracts a rich latent representation of the face that is robust to different views and positions of the face. Our approach is evaluated on several datasets and is shown to outperform existing methods while being robust to uncertainties and image appearances \cite{c15}.

\begin{figure}[!ht]
\centering
\includegraphics[width=3.6in]{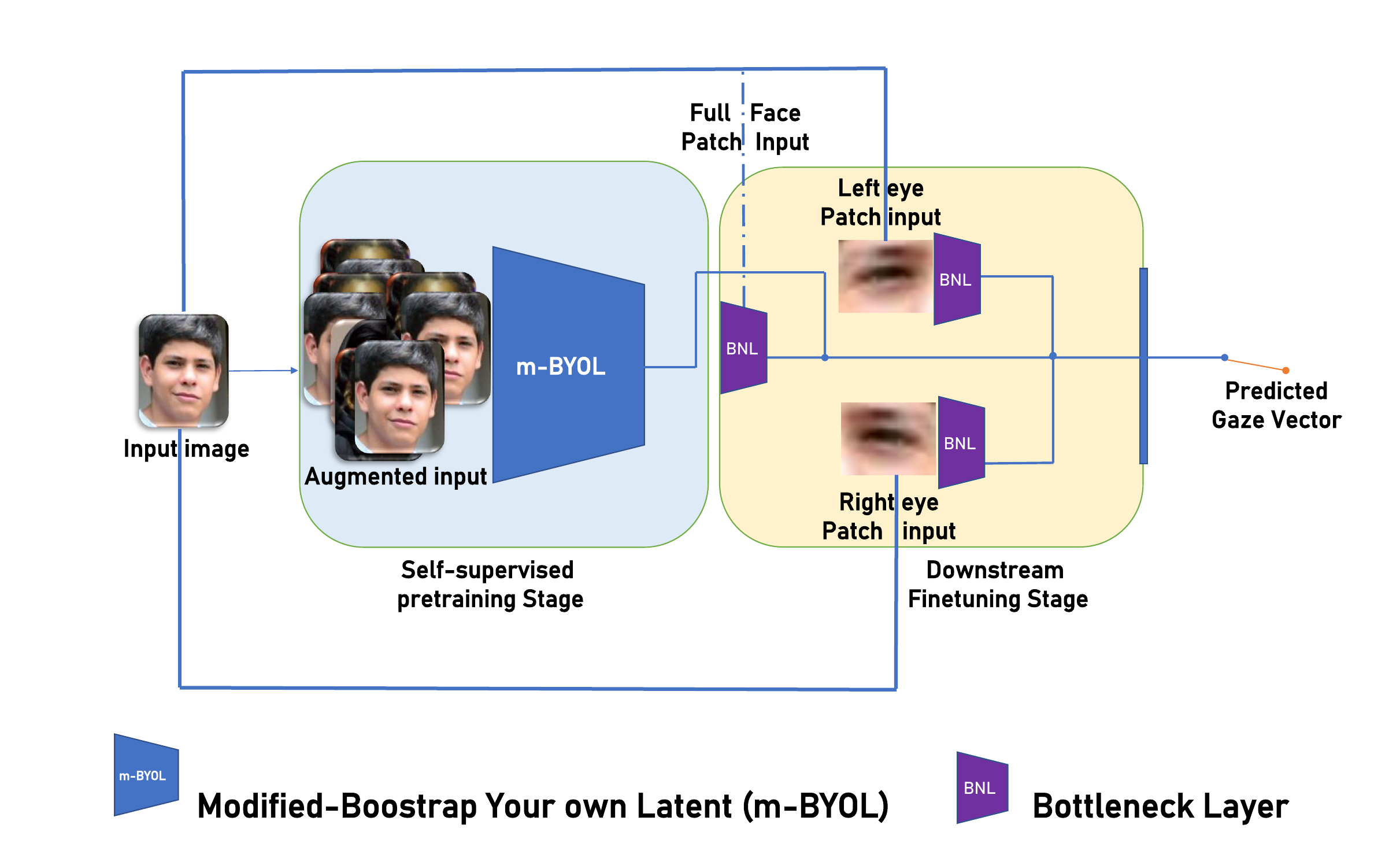}
\caption{\sao{High-Level view of SLYKLatent. The framework consists of two network modules; a self-supervised pretraining module and a patch module network.}}
\label{fig: SLYKLatent Overview}
\end{figure}

\subsection{{Purpose and Contribution} \label{Contribution}}
Since a rich latent embedding can be extracted given a learning function (deep neural network) that solves a set of predefined equations (appearances) on a facial image via self-supervised pretraining \cite{c22}, we hypothesize that the performance of facial cue tasks (gaze estimation in particular)  can be significantly improved if a rich latent representation can be drawn from facial images and individual (left or right) intrinsic eye patch features.
Hence, this study proposes a new learning framework, SLYKLatent, for efficient facial feature estimation, with a specific focus on gaze estimation in the context of human-robot interaction. Our proposed framework addresses the challenges of heteroscedastic noise, domain adaptation, and equivariance shifts in facial feature estimation tasks. The main contributions of this study are:

\begin{itemize}
    \item A framework that utilizes self-supervised learning incorporating a global branch and local branch to extract rich representation, and transfer feature learning to enhance generalizability, uncertainty robustness, noise tolerance, and ability to account for unobserved facial features in facial feature estimation. To introduce appearance equivariance, we incorporate transformations in the self-supervised pretraining stage, which enables the model to learn from diverse appearances in the input data, reducing its reliance on labelled data. The multi-head attention layer allows the model to attend to multiple facial features simultaneously, while the Patch Module Networks (PMN) help extract valuable features from the eyes independently during downstream feature finetuning.

    \item We propose a novel modification to the Bootstrap Your Own Latent (BYOL) self-supervised learning  framework \cite{c17} that includes a new self-supervised loss function. Our modification introduces a new loss function that incorporates the negative cosine similarity between the predictions and feature representations of two different augmented views of the same image. Additionally, we introduce a novel loss function for gaze estimation during downstream fine-tuning that addresses uncertainties in the gaze prediction model by incorporating facial feature weights via explained variance of the gaze features. This approach results in improved accuracy and is a unique contribution to the literature. Our modifications represent a significant step forward in the development of effective deep learning models for gaze estimation and facial expression recognition, demonstrating the potential to improve the performance of deep learning models on these tasks.

    \item A proof of concept of the effectiveness of the proposed framework through the demonstration of state-of-the-art performance on gaze estimation.

    \item A comprehensive set of ablation studies that demonstrate the effectiveness of the components in the proposed framework, as well as an evaluation of results using the explained variance metric.
\end{itemize}

In summary, our proposed framework SLYKLatent offers a solution to the challenges of facial feature estimation by efficiently addressing the problem of achieving domain generalization and robustness to appearance uncertainties while providing improved performance on gaze estimation and facial expression recognition tasks.

\section{Background}
\subsection{Gaze Estimation}

Recent advancements in the field of facial feature estimation have been driven by the availability of large datasets, improvements in deep learning techniques, and a growing interest in utilizing facial cues for various applications \cite{c23, c24}. Facial feature estimation encompasses several tasks, including age estimation, facial recognition, facial detection, and emotion estimation \cite{c25}. In this section, our primary focus is on facial feature estimation within the scope of HRI, with a specific emphasis on gaze estimation.

Facial feature estimation is an important aspect of human-robot interaction, as it enables robots to "understand" and respond to human needs, intentions, and goals. Gaze estimation is a specific task within facial feature estimation that has been widely studied in the context of HRI \cite{c12, c13, c14, c15, c16, c69}. Two distinct methods have been proposed for estimating gaze points from facial images: model-based and appearance-based methods \cite{c26, c27, c28}.

Model-based methods for gaze estimation can be further subdivided into 3D model methods and 2D methods. The 3D model approach regresses gaze points from pre-built person-dependent 3D models of the eyes. However, these approaches are often limited by the need for calibration for each subject, as well as the need for accurate 3D models of the eyes. On the other hand, 2D model methods do not require person-specific calibration, allowing gaze points to be directly regressed from features of 2D models of the eyes \cite{c26, c27, c28}.

Appearance-based methods for gaze estimation can also be subdivided into two categories: the conventional approach and the learning-based approach. Conventional appearance-based approaches regress the human gaze from the raw facial image and generally require algorithms for feature extraction and a regression function that maps extracted features to gaze points. In contrast, the learning-based approach replaces both the feature extraction and regression functions with a deep neural network \cite{c26, c27, c28}.

In recent years, deep learning-based approaches have shown promising results in gaze estimation, as they have proven to be robust to variations in lighting, pose, expression, and other factors that can affect the appearance of the face and the gaze direction \cite{c26, c27, c28, c29}. However, the performance of these methods is still limited by the need for large annotated datasets, as well as the challenges of generalization to unseen data. In the next section, we will review the current state-of-the-art in learning-based appearance gaze estimation, and discuss how these methods can be improved to better handle the challenges of heteroscedastic uncertainty and equivariance shifts.

\subsection{Learning-based Appearance Gaze Estimation}
Given an image, $s$ from a sample space $D$ a learning-based appearance facial cues estimation approach seeks to compute and extract useful deep features from image $s$ to solve a regression or classification task. As with many other vision-based tasks, the learning-based gaze estimation approach often requires a large repertoire of facial image datasets. Based on the type of learning framework used it could also require appropriate annotations. Different learning-based methods have been proposed, but the most popular frameworks have employed supervised learning. These methods often leverage convolutional neural networks (CNNs) to extract visual features from facial images and map them to gaze vectors. Common CNN architectures used in gaze estimation include DenseNet\cite{c66}, ResNet \cite{c65}, and VGGNet \cite{c67}.

The first CNN-based gaze prediction network was introduced in \cite{c68}. They proposed the use of eye patches (feature space) and gaze vector annotation (target label) to map aligned eye features to the corresponding gaze vector. An improved version, GazeNet was later introduced \cite{c33}. This consisted of a CNN-based architecture that aims to learn the low and high-level appearance and map learned features to gaze vector features from a full-face image and corresponding head pose annotation. A Spatial Weight CNN was introduced in \cite{c34} and aims to eliminate noise by allocating more weights to important regions of a face image and low weights to others that contribute to noise, which led to an improvement in the learned target gaze vector. Seeking to improve gaze estimation accuracy and enhance generalizability, Bayesian-enabled CNNs were introduced for gaze estimation \cite{c35}. This approach simply uses ensembles of point-estimate models, such that gaze regression can be carried out on more than a single set of parameters, hence, the framework benefits from several features learned from each of the models. To encode rich visual features extracted by CNNs and at the same time benefit from the temporal details in non-static image data, a combination of a CNN-based network architecture and other Artificial Neural Networks (ANN) was proposed. Examples of these methods include the use of Recurrent Neural Networks (RNNs) and Long Short-term Memory (LSTM) networks \cite{c36, c37, c28, c39}. \sao{To address generalization challenges, approaches like \cite{c73}, which incorporate gaze frontalization within the learning process, have been proposed to enhance domain generalisation.} A comprehensive survey of learning-based appearance gaze estimation methods can be found in \cite{c29}.

\subsection{Gaze Estimation using Facial Patches}
The use of facial patches, particularly eye patches, has been a prevalent approach in gaze estimation research, as the eyes contain rich information about the gaze direction of an individual \cite{c29, c40}. One popular approach for gaze estimation using facial patches is to extract features from the eye patches, concatenate them, and then feed them into a learning function \cite{c14, c29}. \sao{A noteworthy contribution in this area is \cite{c72}, which utilizes multiple low-resolution cameras embedded in the frames of normal glasses to capture different views of the eye and employs learning-based gaze estimation methods to directly regress from eye images to gaze direction.}

Itracker \cite{c40} introduced the concept of fusing features of left, right, and face regions to estimate camera-to-eye distance. In contrast, \cite{c50} used the same approach but to infer gaze vectors from full-face images. Similarly, \cite{c42} trains ensemble networks of eye patches via the sums of supervised losses from the output of each eye network. It is also important to note that while the use of facial patches can improve the performance of gaze estimation models, it is not a complete solution on its own. Other factors such as model architecture, feature extraction techniques, and dataset bias can also greatly influence the overall performance of gaze estimation models.

Additionally, as gaze estimation is a challenging task due to the high variability of head poses, lighting conditions, and facial expressions, it is important to consider the limitations of the data and take into account factors such as appearance uncertainty. Various datasets have been used to train and evaluate gaze estimation models using facial patches. The most commonly used datasets are MPIIGaze \cite{c33}, Gaze360 \cite{c39}, EyeDiap \cite{c58}, \sao{RT-GENE \cite{c42}}  and ETHX-gaze \cite{c53} all of which contain images of people looking in different directions with different lighting conditions, head poses, and facial expressions.

\subsection{Self-supervised Learning for Improving Gaze Estimation}
Despite the vast improvement in the performance of supervised learning, it is usually limited by the requirement for large curated data and corresponding annotations. In the context of face estimation, it is often expensive to set up, collect, clean, and analyze. On the other hand, SSL requires no data annotation nor does it require the complicated setup of a laboratory environment \cite{c21, c22, c45, c46, c47}. Freely available unlabelled image-based data on the internet can be curated, therefore eliminating the heavy lifting of data collection.

SSL has emerged as a prominent alternative to conventional supervised learning in numerous vision tasks, encompassing gaze estimation \cite{c21, c22, c45, c46}. The primary benefit of SSL lies in its elimination of manual data annotation, enhancing cost-effectiveness and efficiency. Furthermore, SSL employs pretext tasks, which are auxiliary tasks designed to learn meaningful data representations without requiring explicit labels \cite{c21, c22, c45, c46}. These representations can subsequently be fine-tuned for downstream applications like gaze estimation.

Recent work has explored the use of SSL for gaze estimation, with a focus on learning latent representations of the full face or eye patches using methods such as Variational AutoEncoders (VAE) and Generative Adversarial Networks (GANs) \cite{c40, c41, c42}. These approaches have demonstrated the effectiveness of SSL in capturing universal features for gaze redirection. However, these methods typically fuse the features of the face and eyes, making it difficult to disentangle the contributions of each. Our approach, SLYKLatent, aims to overcome this limitation by separately extracting features from the full face and eye patches using self-supervised pretraining and then fusing them during downstream finetuning. Other challenges in gaze estimation are the presence of various sources of uncertainty and variations in image appearances. These include, but are not limited to, lighting conditions, head pose, facial expressions, and eye occlusion. These factors can lead to large variations in the appearance of the eyes and face, making it difficult to accurately estimate gaze vectors. To address this challenge, recent work has focused on developing methods that are resistant to these uncertainties and variations in image appearances.

Recently, a method called Swap Affine Transformations (SWAT) has been proposed by \cite{c44}. It builds on the Swapping Assignments between multiple Views of the same image (SwAV) SSL framework \cite{c47} by introducing modifications to the pretext task and augmentation techniques that support equivariance shifts in the input image. While this method has been successful in mitigating the effects of unaccounted-for equivariance properties in gaze estimation, our approach is different in that we use a downstream finetuning technique that takes into account both the features learned from the SSL pretraining process as well as features from the eye patches. Additionally, our SSL framework is based on a modified version of BYOL \cite{c17}, while SWAT is built on the SWAV SSL framework. By combining the features learned from the SSL pretraining with eye patch features, our approach is able to achieve better results than previous methods in terms of gaze estimation accuracy.

Similar to our approach, the work by \cite{c15} proposed the use of an eye patch module network to account for intrinsic eye features in their Modulation Adaptive Eye Guiding Network (MANet) framework. However, there are several key differences between our approach and MANet. Firstly, while they modified the BYOL framework at the representation level, our approach modifies the augmentation view, adds an attention layer of 8 heads, and incorporates an added eye patch network for downstream finetuning. Secondly, our approach achieved a better result when compared with MANet. Lastly, our framework aims to enrich and endow the extracted facial representation with unaccounted appearance properties of the aligned face and features learned from the eye patches, whereas the main focus of MANet is on adapting eye patches for guiding gaze regression.

Previous methods for gaze estimation have used supervised or self-supervised approaches to fuse features from left, right, and face regions or to employ ensemble networks of eye patches. However, these approaches can lead to sub-optimal performance in challenging conditions such as head pose variations and lighting changes. To address these limitations, we propose a novel approach that separately extracts features from the aligned full-face image via a self-supervised pre-trained model and a ResNet encoder, as well as from facial and eye patches via an ensemble of bottleneck layers. By doing so, our approach can better capture the dependencies between eye patches and full-face images and utilize an important visual representation of the full-face image for final gaze regression. Furthermore, we employ self-supervised pretraining to reduce reliance on labelled data and improve performance in the face of appearance uncertainty.

In the following sections, we describe our approach in more detail, including the use of appearance transformations and multi-head attention mechanisms. We also present experimental results to demonstrate the effectiveness of our approach, comparing it with existing methods on the MPIIGaze, Gaze360, and ETHX-gaze datasets, which are commonly used for evaluating gaze estimation models.

\section{The SLYKLatent Framework}
\subsection{mBYOL} \label{Details about mBYOL}
The SLYKLatent framework is designed to be able to tackle the challenges posed by uncertainties, invariance, and positional shifts that may be encountered in novel domains. To do this, it makes use of a modified BYOL \cite{c17} (mBYOL) architecture for self-supervised feature extraction and supervised transfer feature fine-tuning. The mBYOL architecture is a variant of the Bootstrap Your Own Latent (BYOL) architecture, a self-supervised learning framework comprising two parallel but asymmetrical neural networks known as online and target networks. \sao{These two networks are designed to collaborate through a unique mechanism where the online network learns to predict the output of the target network via a slow moving average of the online network's weights \cite{c17}.}

\begin{figure}[!ht]
    \centering
    \includegraphics[width=0.48\textwidth]{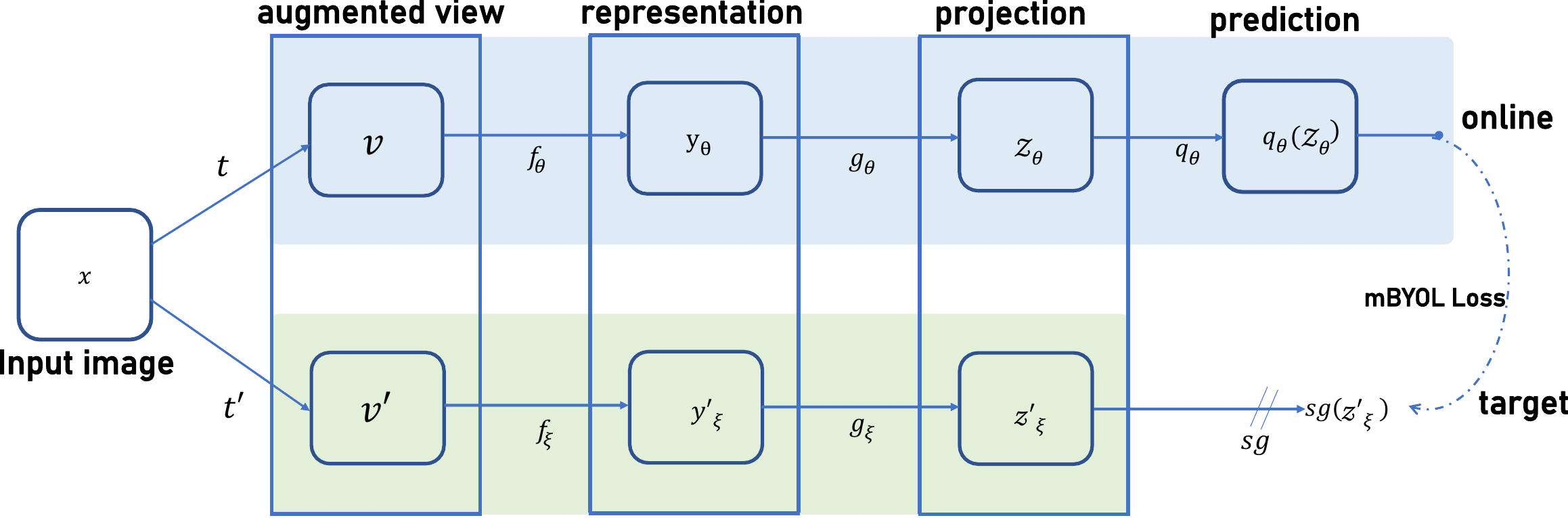}%
    \caption{The Modified Bootstrap Your Own Latent (mBYOL) Architecture. The Architecture is made up of two parallel asymmetrical networks, the target network $\tau'$ and the online network $\tau$. Each network is made up of 3 essential stages: augmented view $v$; representation stage $y$ (this is where image embeddings are computed), and; projection layer $z$. In addition to this, the online network includes a prediction stage. The mBYOL loss computes the Negative Cosine similarities between the online and target network.}
    \label{fig: BYOL Architecture}
\end{figure}

To create the mBYOL framework, we made some changes to the BYOL architecture, in the augmentation view section, the representation layer, and the loss computations. 

\subsubsection{Augmentation View}
\begin{figure}[!ht]
    \centering
    \includegraphics[width=.47\textwidth]{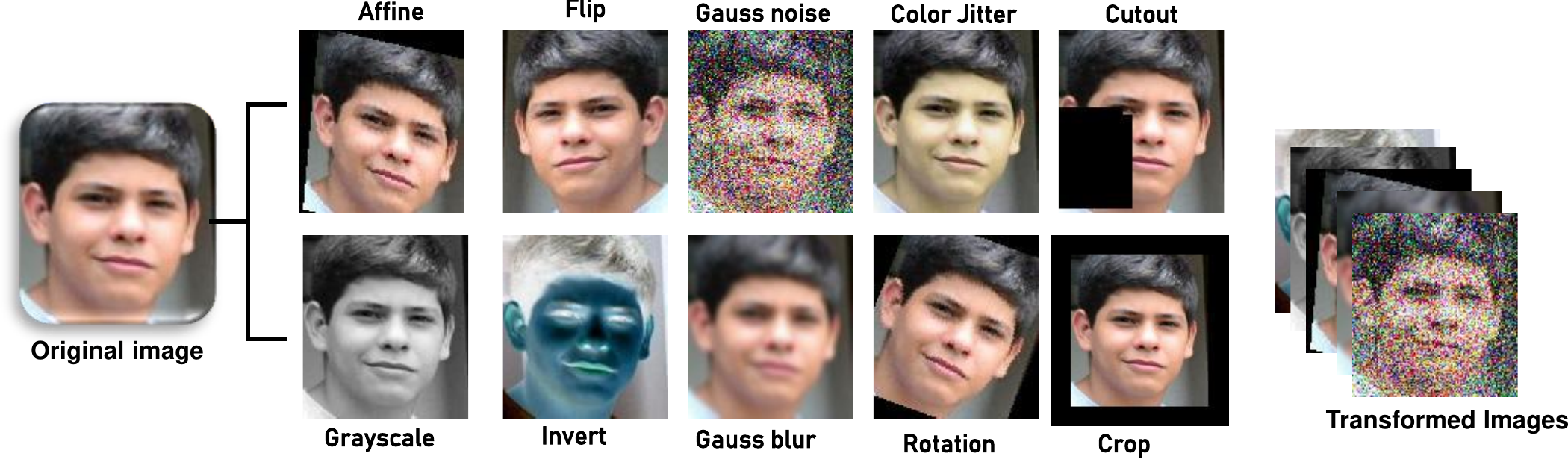}%
    \caption{The Augmentation view applied on mBYOL.}
    \label{fig:Augmentation view}
\end{figure}
We adjusted the augmented view by randomly applying a set of transformation sets in a specific order. Each transformation has a probability value $p$ that determines if it will be included in the data samples for a given batch during training. The transformation sets include a range of techniques such as Random Horizontal Flip, Gaussian Blur, Random Affine, Random Rotation, Random Crop, Center Crop, Random Grayscale, Color Jitter, Random Invert, Gaussian Noise, and Cutout. The applied modifications to the augmented view help to address the issues arising from appearance uncertainties, equivariance, and positional shifts. For example, the Random Affine and Random Rotation transformations are effective in addressing positional shifts and equivariance, while Gaussian Blur and Random Grayscale are beneficial in mitigating appearance uncertainties. By addressing these challenges, the modified architecture is better equipped to tackle the challenges of downstream finetuning, leading to an overall improvement in performance. 

\subsubsection {Modification of the Representation layer} \label{Modification Representation Layer}
We introduced a novel, local-global architecture into the mBYOL framework. The modified model includes two separate local branches that independently process the left and right eye images of a face. 
\sao{We extracted the left and right eye images using dlib facial landmark detector \cite{c74}, resizing them to $36$ x $60$ pixels, without using gaze labels or other supervised information. This approach preserves the self-supervised nature of the mBYOL framework in the local branch.} Each local branch consists of three Convolutional layers with 32, 64, and 128 filters, respectively, each followed by batch normalization and ReLU activation \cite{c61}. A global average pooling layer and a fully connected layer with 52 outputs are applied to the output of the last Convolutional layer. These local branches work in parallel to extract local features from each eye.

The global branch uses a configurable backbone Inception-ResNet-V2 \cite{c63} model pre-trained on ImageNet. The global branch processes the global face image and outputs a feature vector.

To emphasize the importance of different regions in the global and local features, we also introduce multi-head attention mechanisms to each branch. An 8-head attention layer is applied to the output of the global backbone and to each of the local branches, \sao{a decision driven by the need to capture a broad spectrum of features from high-dimensional facial data effectively. Employing multi-head attention helps in parallel processing of the aforementioned features, allowing the model to attend to different aspects of the input data simultaneously. The choice of 8 heads helps to balance computational efficiency with performance and aligns with findings from previous research that demonstrates the efficacy of this configuration in similar tasks involving complex visual processing \cite{c70}.}

Finally, the features from the local and global branches are concatenated to form a combined representation. This representation then passes through the projection in both the online and target networks. The online network further processes the representation through a prediction head, similar to the original BYOL framework \cite{c17}. This modified architecture, therefore, learns rich and diverse representations by capturing both the local eye-specific features and the global face features in a self-supervised manner, which leads to improved performance in downstream tasks, as will be detailed next.

\subsubsection{Architecture details of mBYOL}
\label{Architecture details of mBYOL}
In the mBYOL architecture, the encoder $f_\theta$ is formed by a combination of global and local branches as detailed in Section \ref{Modification Representation Layer}. The concatenated features from these branches are then passed through the projection head, a Multi-Layer Perceptron (MLP) with dimensions (1536, 1024, 1024), using ReLU activation \cite{c61}. This head refines the representation for the task of predicting the target view in the contrastive learning setup.

\textbf{Online Network:} The online network encompasses the complete architecture described above, including the global and local branches, multi-head attention, and the projection and prediction heads. Following the projection head, the prediction head, with dimensions (1024, 1024, 1024), further processes the representation, shaping the learning objective. The online network's weights are actively trained during the learning process, and it processes two augmented views $v$ and $v'$, extracting representations as in (\ref{eqn: representation for both views}):

\begin{equation}
\label{eqn: representation for both views}
   {y_\theta} =  (y_{global} \oplus y_{left} \oplus y_{right})
\end{equation}
for both views. Where $\oplus$ represents the concatenation process.

\textbf{Target Network:} The target network mirrors the online network but excludes the prediction head \cite{c17}. It also processes $v$ and $v'$ creating projections for both views. The parameters of the target network follow the parameters of the online network with a delay, using an exponential moving average (EMA) controlled by a momentum term \cite{c17}, as detailed in (\ref{eqn:paramter update}) using a target decay rate $\mathbf{\tau}$ $\mathbf{\epsilon}$ ${[0, 1]}$. 

\begin{equation}\label{eqn:paramter update}
    \centering
    \xi \leftarrow \tau \xi + (1 - \tau) \theta
\end{equation}
The target network's role is to generate stable representations that the online network's predictions aim to match, contributing to a more stable and effective learning process. 

The architecture also includes momentum updates for various components, including the backbone, local branches, and multi-head attention layers, and utilizes the negative cosine similarity loss function \cite{c17}. The regression loss $\mathcal{L}_{SSL}$, defined as: 

\sao{
\begin{align}
\label{eqn:SSL_loss}
\mathcal{L}_{\text{{SSL}}} = \frac{1}{4} \Bigl( &\text{{NS}}_{\text{{cos}}}(q_{\theta}(v), z_{\xi}(v')) \nonumber \\
+ &\text{{NS}}_{\text{{cos}}}(q_{\theta}(v'), z_{\xi}(v)) \nonumber \\
+ &\text{{NS}}_{\text{{cos}}}(z_{\theta}(v), z_{\xi}(v')) \nonumber \\
+ &\text{{NS}}_{\text{{cos}}}(z_{\theta}(v'), z_{\xi}(v)) \Bigr)
\end{align}
}
is minimized through stochastic optimization. In this expression, \( q_{\theta}(v) \) and \( z_{\theta}(v) \) are the prediction and projection of the online network for view \( v \). \( z_{\xi}(v') \) is the projection of the target network for view \( v' \). \(\text{{NS}}_{\text{{cos}}} \) represents the negative cosine similarity loss function.
This loss function aligns the representations of the augmented views \( v \) and \( v' \), encouraging the online network to predict the target network's projections, and vice versa, thereby enhancing the learning of meaningful representations.

\begin{figure*}[!ht]
    \centering
    \includegraphics[width=\textwidth]{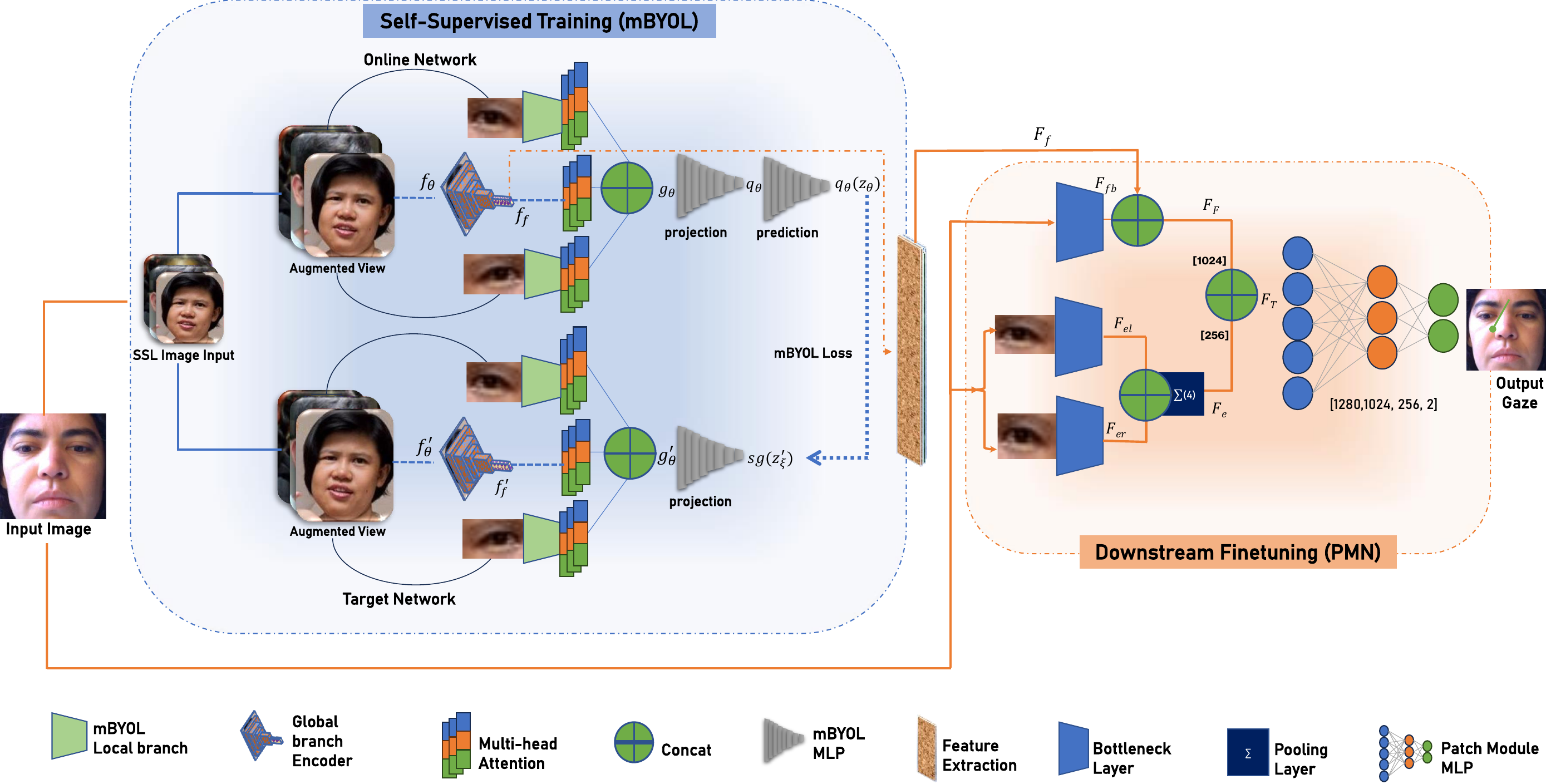}%
    \caption{Schematic diagram of the proposed framework, SLYKLatent. SLYKLatent comprises mBYOL, a modification of BYOL framework\cite{c17}; and a downstream finetunning which is made up of an eye-patch module network. The resultant features are concatenated to regress against the ground-truth gaze vector. }
    \label{fig: SLYKLatent Architecture}
\end{figure*}%

$\mathcal{L}_{SSL}$ compares the normalized version of the predictor (online network), and the projector (target network), calculated after every training step. After the specified training epochs are completed the encoder $f_\theta$ is saved and all the other parameters are discarded. Further details on training and inference are presented in Section \ref{sec:implementation details}. 

This cohesive architecture, along with the collaboration between the online and target networks, enables a robust self-supervised learning process. By aligning the features of various parts of the input image and transforming them through carefully designed network components, mBYOL provides a versatile and effective framework for learning rich representations from complex and high-dimensional data without the need for labeled supervision.
\subsection{Downstream Transfer Learning Fine-tuning}
Once the self-supervised pretraining process is complete, the SLYKLatent framework progresses to the downstream fine-tuning stage. This stage diverges from traditional approaches that rely on linear evaluation and regression of image features after self-supervised learning pretraining. Instead, the SLYKLatent framework introduces the concept of Patch Module Networks (PMN) specifically designed for gaze estimation. The PMN consists of bottleneck modules that are instrumental in extracting relevant features from the face and eye patches, thereby facilitating the training of a precise supervised gaze estimation. \sao{During this stage, the pre-trained mBYOL model, initially trained using self-supervised methods, is further adapted to the task of gaze estimation. Unlike conventional fine-tuning approaches where the pretrained model is frozen to retain learned features, our approach continues to train the mBYOL component. This strategy ensures that the model retains the generalizability learned from the diverse facial expression dataset and also adapts these features to enhance performance on specific-gaze-related tasks. By allowing the mBYOL model to update during this phase, we ensure a more seamless integration of learned representations, making the feature set more adaptable and relevant to the downstream task. This approach effectively combines the benefits of self-supervised learning with the specificity of supervised fine-tuning, promoting a robust adaptation to the gaze estimation task.}
During this stage, the mBYOL model is used as a backbone to extract features from the face and set to evaluation mode with additional features learned from the PMN. Comprising three specific bottleneck modules—the face bottleneck \(B_{f}\), left eye bottleneck \(B_{l}\), and right eye bottleneck \(B_{r}\)—the PMN processes distinct facial regions for feature extraction and transformation.

The architecture of the bottleneck modules, as detailed in Table \ref{tab:bottleneck_pmn}, consists of three convolutional layers each followed by a batch normalization layer. The output of the final convolutional layer is passed through an adaptive average pooling layer, reducing the spatial dimensions while preserving the channel information, and finally through a fully connected layer to produce a 512-dimensional feature vector.

\begin{table}[htbp]
\caption{Bottleneck Module Architecture}
\label{tab:bottleneck_pmn}
\centering
\resizebox{\columnwidth}{!}{%
\begin{tabular}{lclcc}
    \hline
    Layer & Type & Output Channels & Kernel Size & Padding \\
    \hline
    Conv1 & Conv2d & 32 & 3 & 1 \\
    BN1 & BatchNorm2d & 32 & - & - \\
    Conv2 & Conv2d & 64 & 3 & 1 \\
    BN2 & BatchNorm2d & 64 & - & - \\
    Conv3 & Conv2d & 128 & 3 & 1 \\
    BN3 & BatchNorm2d & 128 & - & - \\
    Pooling & AdaptiveAvgPool2d & 128 & 1x1 & - \\
    FC & Linear & 512 & - & - \\
    \hline
\end{tabular} %
}
\end{table}

These bottleneck modules serve multiple purposes in the SLYKLatent framework. They reduce the dimensionality of the input features, making it easier to train the downstream gaze estimation model, and provide a way to extract relevant features from the face and eyes. In addition to reducing dimensionality, the bottleneck modules introduce non-linearity into the network, allowing it to learn more complex features specific to the task of gaze estimation. The face and eye bottleneck modules follow the same architecture and play a crucial role in capturing the most important features of the input image. While the face bottleneck is focused on capturing global face features, the eye bottlenecks are tailored for extracting detailed eye-specific information. Together, these components enable the SLYKLatent framework to learn rich and robust representations that can be used to accurately estimate gaze vectors in various scenarios.

\subsection{SLYKLatent Downstream Gaze Estimation}
\label{subsec:DownstreamFeatureFinetuning}
In the SLYKLatent framework, downstream gaze estimation is performed after feature extraction using the SSL encoder and PMN. The feature vectors produced from the SSL encoder and PMN are then fed into a MLP that predicts the gaze vector.

The process begins with the generation of normalized landmarks from the full-face image \(X\), which is used to extract the eye patches \sao{using the techniques employed in \cite{c42}}, denoted as \(X_{el}\) and \(X_{er}\) for the left and right eyes, respectively. \sao{The dimension of the eye patches is then resized to a size of $36$ x $60$}, these eye patches are input into the corresponding eye bottleneck modules of the PMN, producing feature vectors \(F_{el}\) and \(F_{er}\). These feature vectors are then concatenated and an average pooling operation is performed, resulting in a reduced-dimensionality eye feature vector \(F_{e}\). Mathematically, this can be represented as:

\begin{equation}
\label{eqn:Eye-Features}
    {F}_{e} = {Pool_{avg}}(F_{el} \oplus F_{er})
\end{equation}

Simultaneously, the full-face image \(X\) is processed through the SSL encoder \(F_{\theta}\) and the face bottleneck module to extract the facial feature vectors \(F_{f}\) and \(F_{fb}\), respectively. These vectors are then combined to form a comprehensive face feature vector \(F_{F}\) as follows:

\begin{equation}
\label{eqn: Face Concatenation}
    F_{F} = F_{f} \oplus F_{fb}
\end{equation}

The comprehensive face feature vector \(F_{F}\) and the reduced-dimensionality eye feature vector \(F_{e}\) are then concatenated to form the final feature vector \(F_{T}\) that is passed into the MLP:
\begin{equation}
\label{eqn: Face-Eye Concat}
    F_{T} = F_{F} \oplus F_{e}
\end{equation}

The MLP, denoted by \({MLP}\), varies based on the dataset being used. For the MPIIFace dataset, the MLP is composed of three layers. The initial layer transforms the input feature vector of size 1280 to a size of 1024, followed by batch normalization. Subsequent layers further reduce the dimensionality to 256, and finally to 2, which represents the gaze vector. Dropout regularization with a rate determined by the dataset-specific configuration is applied in the architecture. As the range of the gaze vector in the MPIIFace dataset is between -1 and 1, the $tanh$ activation function is utilized to ensure the output values fall within this range:

\begin{equation}
    \label{eqn:MLP-MPII}
    {MLP}_{MPII}: 1280 \rightarrow 1024 \rightarrow 256 \rightarrow 2
\end{equation}

For other datasets, the MLP follows a similar architecture but without the dropout regularization and the $tanh$ activation function in the final layer:
\begin{equation}
    \label{eqn:MLP-Other}
    {MLP}_{MPII}: 1280 \rightarrow 1024 \rightarrow 256 \rightarrow 2
\end{equation}

The $MLP$ predicts the gaze vector $\hat{y}$ as:

\begin{equation}
\label{eqn: Gaze-vector output}
    \hat{y} = {MLP}(F_{T})
\end{equation}

The output of the MLP is a 2-dimensional vector representing the predicted pitch and yaw angles, that is, $\hat{y}=(\theta_{pitch},\theta_{yaw})$. 
These angles are defined as follows:

\begin{itemize}
\item $\theta_{pitch}$ ranges from $\SI{-90}{\degree}$ to $\SI{90}{\degree}$, with positive values indicating an upward gaze direction.
\item $\theta_{yaw}$ ranges from $\SI{-180}{\degree}$ to $\SI{180}{\degree}$, with positive values indicating a clockwise rotation of the gaze direction when looking down from above.
\end{itemize}

These angles are subsequently transformed into a normalized 3D gaze direction vector. This conversion is a critical post-processing step in the gaze estimation pipeline, ensuring that the predicted gaze direction can be represented in a 3D space for accurate angular comparisons. The conversion process from angular outputs to 3D Cartesian coordinates is formalized as:

\begin{equation}
\begin{aligned}
x &= \cos(\theta_{pitch})  \sin(\theta_{yaw}) \\
y &= \sin(\theta_{pitch}) \\
z &= \cos(\theta_{pitch}) \cos(\theta_{yaw})
\end{aligned}
\end{equation}
where $x$, $y$, and $z$ represent the Cartesian components of the resulting unit length vector $v = (x, y, z)$.

In this representation:
\begin{itemize}
    \item The $x$-coordinate corresponds to the lateral direction, with positive values denoting rightward from the perspective of the participant.
    \item The $y$-coordinate corresponds to the vertical direction, with positive values denoting upward from the perspective of the participant.
    \item The $z$-coordinate corresponds to the frontal direction, with positive values denoting forward, away from the participant.
\end{itemize}
This normalized gaze direction is utilized in the computation of the gaze angular error in the subsequent performance evaluation.

During training, the gaze estimation loss \(\mathcal{L}_{SUP}\) is computed to quantify the difference between the predicted gaze vectors \(\hat{y}\) and the actual gaze vectors $y$. The loss function incorporates a weighting scheme based on the inverse of the explained variance, \(\omega\). This strategy emphasizes the minimization of larger errors, resulting in the model learning to accurately predict gaze vectors while concurrently penalizing larger errors more heavily than smaller ones.

The Mean Absolute Error (MAE) between the predicted and actual gaze vectors for each sample in a batch of size \( n \) is calculated as follows:

\begin{equation}
\label{eqn: Mean Absolute Error}
\mathcal{L} = \frac{1}{n} \sum_{i=1}^{n} ||{y}_i - \hat{y}_i||
\end{equation}
where \( y_i \) and \( \hat{y}_i \) denote the actual and predicted gaze vectors for the \( i \)th sample, respectively.

The Total Sum of Squares (SST), which measures the total variance in the gaze vectors, and the Sum of Squared Errors (SSE), which measures the total deviation of the predicted gaze vectors from the actual gaze vectors, are calculated as follows:
\begin{equation}
    \label{eqn: Total Sum of Squares}
    \text{SST} = \sum_{i=1}^{n} ||{y}_i - \bar{y}||^2
\end{equation}

\begin{equation}
\label{eqn: Sum of Squared Errors}
    \text{SSE} = \sum_{i=1}^{n} ||{y}_i - \hat{y}_i||^2
\end{equation}
where \( \bar{y} \) denotes the mean of the actual gaze vectors. The explained variance \( V_{EX} \), which measures the proportion of the gaze vector variance that is predictable from the predicted gaze vectors, is then calculated as:
\begin{equation}
    \label{eqn: Explained Variance}
    V_{EX} = 1 - \frac{\text{SSE}}{\text{SST}}
\end{equation}
The weight \( \omega \) for each sample, which corresponds to the inverse of the explained variance, is computed as:
\begin{equation}
    \label{eqn: Inverse Explained Variance}
    \omega = 1 - V_{EX} = \frac{\text{SSE}}{\text{SST}}
\end{equation}

Finally, the gaze loss metric \( \mathcal{L}_{SUP} \), which quantifies the difference between the predicted and actual gaze vectors across the entire batch, is calculated as the product of the MAE and the factor \( \omega + 1 \):
\begin{equation}
    \label{eqn: Supervised Loss}
    \mathcal{L}_{SUP} = \mathcal{L} \cdot (\omega + 1)
\end{equation}

By using this weighted loss function, the model prioritizes the minimization of larger gaze estimation errors, which contribute more significantly to the overall loss.

\section{Experiments and Methodology}
\subsection{Datasets}\label{subsec:Datasets used}
For SSL pre-training, we used the AFFECTNet Dataset \cite{c52}, and for downstream evaluation, we employed MPIIFaceGaze \cite{c34}, Gaze360 \cite{c39}, and ETHX-Gaze \cite{c53}. Detailed descriptions of these datasets, including their composition, collection methodologies, and specific usage in our study, are provided in the Supplementary Material.
We utilized these datasets to evaluate our approach and compare its performance against existing state-of-the-art models in both gaze estimation and facial expression recognition tasks. The outcomes of these assessments are presented and discussed in the subsequent sections of this paper.

\subsection{Implementation details}\label{sec:implementation details}

\textbf{Training details for SSL:}  For self-supervised pretraining, the Stochastic Gradient Descent optimizer \cite{c62} was used with a learning rate of 0.06 and a batch size of 112. The number of epochs was set to 100. EMA was initially set to 0.996 and subsequently increased to 1 during training \cite{c17}. For optimal efficiency and speed during training, the data loading was parallelized across 4 Nvidia A100 GPUs using a distributed data-parallel technique. Training took a total of 80 hours. All model development, training and evaluation work was undertaken using the Pytorch Lightning deep-learning framework \cite{c56}.

\textbf{Training details for Gaze estimation:} For gaze estimation downstream fine-tuning, the Adam optimizer \cite{c71} was used for all datasets with different batch sizes and learning rates for each of the datasets. We used a learning rate of 0.0003 and a batch size of 16 for all gaze estimation datasets. As with SSL training, a distributed data-parallel implementation was adopted for data loading during the training phase. 
To adapt the model to the specific characteristics of each dataset, we employed dataset-specific configurations. For every dataset, a learning rate annealing strategy was adopted, where the learning rate was reduced upon observing plateaus in training/validation performance. This dynamic adjustment of the learning rate aids in achieving better convergence and prevents overshooting in the optimization landscape.

For the Gaze360 dataset, the model was trained for a total of 10 epochs. The primary metric for monitoring the learning rate scheduler, in this case, was the validation angular error. Training took 4 hours on 4 x v100 Nvidia GPUs.

In our experimentation with the MPIIFace dataset, we emulated the original work's methodology, implementing leave-one-person-out cross-validation \cite{c34}. This approach rigorously assessed the model's efficacy. The training spanned a maximum of 30 epochs for each participant. 
To avoid over-fitting early stopping was employed with a patience of 2 epochs and a minimum delta of 0.1 for the angular error. Post-training, 15 distinct models, each corresponding to left-out participants, underwent testing. The resulting angular errors were averaged to compute a mean angular error. Training took 17 hours on 4 x V100 Nvidia GPUs.

For the ETH dataset, the validation angular error served as the primary metric for the learning rate scheduler. Callbacks were strategically placed to monitor model checkpoints based on this error, facilitating the saving of optimal model weights. Additionally, an early stopping mechanism was integrated, endowed with a patience of 20 epochs and a minimum delta of 0.01. Training took 6 hours on 2 x A100 Nvidia GPUs.

In summary, our training methodology, underpinned by the Adam optimizer, dynamic learning rate adjustments, and early stopping, ensures rigorous, robust, and efficient model training tailored to each dataset's nuances.

\section{Results}
\subsection{Performance Evaluation} \label{Results}
In this section we present results comparing SLYKLatent with existing and current
state-of-the-art methods in gaze estimation for each of the gaze datasets presented in Section \ref{subsec:Datasets used}. We also present the results of ablation studies on several components of SLYKLatent to highlight the contribution of each component. As detailed in the Section \ref{subsec:DownstreamFeatureFinetuning}, the predicted 2D angles from the MLP are converted to 3D unit vectors. Similarly, the ground truth data is also represented as 3D unit vectors. This ensures consistent and accurate representation for comparing the predicted and actual gaze directions. The traditional gaze angular error metric,  $\mathcal{L}_{ang}$, is chosen as the performance metric. This metric quantifies the angular discrepancy between the ground truth and the predicted gaze vectors in a 3D space. This is defined as

\begin{equation} \label{eqn:Gaze Angular Error}
     \mathcal{L}_{ang} = \frac{180}{\pi} \cos^{-1}\left( v \cdot  \hat{v} \right)
\end{equation}
where $v$ represents the ground truth 3D gaze direction, and $\hat{v}$ represents the estimated direction.

\subsubsection{Performance of SLYKLatent on Gaze Datasets}
Table \ref{tab:SoTA comparison} provides an exhaustive evaluation of the performance of SLYKLatent, our proposed framework, against existing state-of-the-art methodologies for the MPIIFaceGaze, ETHX-Gaze, and Gaze360 datasets. Notably, SLYKLatent achieves superior performance to existing state-of-the-art models for the Gaze360 dataset. It registers an angular error of $\SI{10.70}{\degree}$ for the full Gaze360 dataset, $\SI{10.34}{\degree}$ for data within $\pm \SI{90}{\degree}$, and $\SI{8.03}{\degree}$ for data within $\pm \SI{20}{\degree}$, surpassing previous benchmarks \cite{c59} \cite{c15} by between $1\%$ and $8\%$ for data within these angular ranges.
For the MPIIFaceGaze dataset, SLYKLatent achieves an angular error of $\SI{3.77}{\degree}$, which is on a better than the best performing method, L2CS-Net \cite{c59} ($\mathcal{L}_{ang}=\SI{3.92}{\degree}$), and superior to other contemporary self-supervised gaze estimation methods such as MANet \cite{c15} ($\mathcal{L}_{ang}=\SI{4.30}{\degree}$) and SWAT \cite{c44} ($\mathcal{L}_{ang}=\SI{5.0}{\degree}$). L2CS-Net is a supervised approach that leverages a CNN architecture and combines loss functions for both the regression of the gaze vector and the classification of varying gaze angles.

\begin{table}[!h]
\caption{Comparison of the mean gaze angular error ($\mathcal{L}_{ang}$) performance of SLYKLatent and alternative approaches in the literature for selected benchmark datasets (MP: MPIIFace, G360: Gaze360, ETH: ETH-XGaze). The best results are highlighted in \textcolor{blue}{blue} and the second best results are in \textcolor{brown}{brown}. The second column is the number of model parameters in millions (M).}

\label{tab:SoTA comparison}
\begin{adjustbox}{width=\columnwidth,center}
\begin{tabular}{lccccccc}
\hline
\multirow{1}{*}{\textbf{Method}}     & \multirow{1}{*}{\textbf{Params}} & \multirow{1}{*}{\textbf{Init Arch.}}       & \multirow{1}{*}{\textbf{MP}}            & \multirow{1}{*}{\textbf{G360}}        & \multirow{1}{*}{\textbf{G360}}    &    \multirow{1}{*}{\textbf{G360}}     & \multirow{1}{*}{\textbf{ETH}} \\
                            & (M)                              &                                   &                                & ($\pm \SI{180}{\degree}$)   &($\pm \SI{90}{\degree}$) & ($\pm \SI{20}{\degree}$)                            &     \\
\hline
GazeNet \cite{c33}          & -                           & ResNet-50                         & -                                 & $\SI{12.8}{\degree}$            & -                             & -        & -\\
EyeDiap  \cite{c58}         & -                            & ResNet-50                         & -                                 & $\SI{12.8}{\degree}$            & -                             & -        & -\\
Full-Face \cite{c34}        & -                           & AlexNet+SW                        & $\SI{4.9}{\degree}$                 & $\SI{15.0}{\degree}$            & -             & -        & -\\
Dilated-Net \cite{c60}      & -                           & Dilated-CNN                       & $\SI{4.8}{\degree}$                 & -                             & -                           & -        & -\\
Gaze360 \cite{c39}          & 24.0                            & Pinball LSTM                      & $\SI{12.1}{\degree}$                & $\SI{13.2}{\degree}$            &$\SI{11.4}{\degree}$             &  $\SI{11.1}{\degree}$       & - \\
ETH-XGaze \cite{c53}        & 23.5                            & ResNet-50                         & $\SI{4.8}{\degree}$     & -                       & -         & -         & $\SI{4.5}{\degree}$ \\
MANet \cite{c15}            & 29.5                            & ResNet-18                         & $\SI{4.3}{\degree}$                 & $\SI{13.2}{\degree}$            & -                             & -        & - \\
SWAT \cite{c44}             & -                            & ResNet-50                         & $\SI{5.0}{\degree}$                 & \textcolor{brown}{$\SI{11.6}{\degree}$}        & -              & -        &$\SI{4.4}{\degree}$\\
L2CS-Net \cite{c59}         & 21.2                            & ResNet-50                         & $\textcolor{brown}{\SI{3.92}{\degree}}$     & -                    & $\textcolor{brown}{\SI{10.41}{\degree}}$ &$\textcolor{brown}{\SI{9.02}{\degree}}$  &- \\
\hline
SLYKLatent                  & 25.5                            & Inception-V1                      & $\textcolor{blue}{\SI{3.77}{\degree}}$     & $\textcolor{blue}{\mathbf{\SI{10.70}{\degree}}}$      & $\textcolor{blue}{\mathbf{\SI{10.34}{\degree}}}$ &$\textcolor{blue}{\mathbf{\SI{8.03}{\degree}}}$       & $\textcolor{blue}{\mathbf{\SI{3.89}{\degree}}}$ \\
\hline
\end{tabular}
\end{adjustbox}
\end{table}

An objective comparison of our model's performance with the ETHX-Gaze dataset was not feasible as SLYKLatent requires the presence of eye patches to perform inference - a requirement not shared by some of the alternative methods in the literature. Consequently, we had to exclude ETHX-Gaze samples where detection of eye patches was not possible. This represented 65\% of the dataset. For the remaining 35\% of samples SLYKLatent achieves an angular error of $\SI{3.89}{\degree}$.

\subsection{Ablation Studies} \label{sec:Ablation Studies}
In this section, we analyze the performance of SLYKLatent through a series of ablation studies, where different components of the model are removed or altered to understand their impact on overall performance. Additionally, we compare the performance of the SLYKLatent model with one that uses BYOL \cite{c17} as its backbone.  Specifically, Table \ref{Tab: Ablation Results} presents results for the following ablated configurations: 
\begin{enumerate}
    \item without the PMN (w/o-PMN) 
\item without SSL (w/o-SSL)
\item without inverse explained variance (w/o-inv-EV)
\item without the mBYOL modification (w/o-mBYOL)
\item basic BYOL \cite{c17} without any enhancements
\sao{\item without the global branch (wo-global)
\item without the local branch (wo-local)}
\end{enumerate}
%
\sao{\textbf{Rationale for Ablation Study Design: } In the design of our ablation studies, we emphasized modifications that directly assess the enhancements made to the traditional BYOL architecture, specifically the introduction of local and global branches. We reference the original BYOL study which highlights the important interaction between the online and target networks \cite{c17}. These networks are foundational to the architecture, as they prevent feature collapse by maintaining a dynamic, yet stable learning target, which is crucial for the robust learning of features without using negative pairs \cite{c17}. This approach allowed us to efficiently utilize our computational resources, focusing on the most impactful elements of our model architecture for the task-specific demands of gaze estimation.
}

The performance metric used is the gaze angular error evaluated on the validation/test data for each dataset. 

\textbf{Impact of Patch Module Network}: We evaluated the contribution of the PMN by analyzing the performance of an SSL-only variant. This model variant included the mBYOL backbone and inv-EV cost function but not the other enhancements of the PMN. The results indicate that the SSL-only version performs reasonably well but is consistently surpassed by the complete SLYKLatent model. The average impact on performance across the datasets is $7.86\%$. This confirms the significance of the PMN in conjunction with mBYOL in providing a more balanced and effective solution for gaze direction prediction.

\textbf{Impact of SSL pretraining}: To assess the impact of the SSL pre-trained model, we compared the results without SSL pretraining (w/o-SSL) with the full SLYKLatent architecture. We eliminated the self-supervision stage of SLYKLatent and replaced the mBYOL pre-trained model with the bare backbone (InceptionResNet-V1) and retained the downstream finetuning PMN. The resulting architecture was then trained using the same experimental setup as used with SLYKLatent (see Section \ref{sec:implementation details}).  The results highlight the benefit of SSL with the angular error increasing by $6.31\%$ on average across the datasets when it is omitted.

\textbf{Impact of Inverse Explained Variance}: To isolate the effect of inv-EV, we trained and tested a model without this modification (w/o-inv-EV), keeping all other aspects and hyperparameters unchanged.  This resulted in an average increase in angular error of $13.01\%$ over SLYKLatent, confirming the  positive effect inv-EV  has on model performance.

\textbf{Impact of mBYOL}: To evaluate the impact of employing our modified BYOL (mBYOL) backbone in the SLYKLatent model, we contrasted its performance with a variant that employs the standard BYOL backbone instead (denoted as w/o-mBYOL). Training was undertaken as described in Section \ref{sec:implementation details}. An examination of the results in Table \ref{Tab: Ablation Results} reveals that w/o-mBYOL is outperformed by SLYKLatent on all three datasets,  with its $\mathcal{L}_{ang}$ on average $8.41 \%$ greater than SLYKLatent.  Notably, while inferior to SLYKLatent,  w/o-mBYOL outperforms the w/o-SSL variant, indicating that the inclusion of BYOL assists the model in learning better representations. Overall, the results indicate that integrating mBYOL as the backbone in the SLYKLatent architecture yields superior performance compared to the standard BYOL backbone.

\sao{\textbf{Impact of Local and Global Branches:} In addition to the above ablations, we evaluated the contributions of the local and global branches of mBYOL by removing each component individually and training the self-supervised models using the same hyperparameters as SLYKLatent. The resulting SSL models (wo-local and wo-global) were then used as backbones for individual downstream fine-tuning on each dataset. The percentage increase in angular error for these ablations, compared to the full SLYKLatent model, is shown in Table \ref{Tab: Ablation Results}.

When the local branch was removed (wo-local), the model experienced a substantial increase in angular error, with an overall degradation of up to 84.23\%. This highlights the significant role played by the local branch in fine-grained gaze estimation, particularly when dealing with larger angular ranges and more complex datasets. In contrast, the global branch appears to contribute more modestly, as removing it (wo-global) resulted in an increase in angular error of up to 4.3\%. While the global branch still plays an important role in improving the model's performance, its absence has a less drastic impact compared to the local branch.

These findings confirm that the local branch is essential for achieving high precision in gaze estimation, especially in scenarios involving a wide range of gaze angles. The global branch, while helpful, is not as critical, but its contribution to overall performance should not be overlooked.}

\textbf{Impact of all enhancements}: To further demonstrate the combined benefit of the different components of SLYKLatent, we trained an SSL backbone using only the BYOL framework, that is, without the proposed mBYOL modification and without the PMN and inv-EV cost function enhancements. This `Basic BYOL' variant yielded an angular error of \SI{4.56}{\degree}, \SI{13.56}{\degree}, and \SI{5.0}{\degree} on the MPIIFace, Gaze360, and ETHX-Gaze datasets, respectively. This corresponds to an average increase in $\mathcal{L}_{ang}$ of $28.5\%$ compared to SLYKLatent. The scale of this increase underscores the accumulative benefit of the PMN, inv-EV and mBYOL enhancements.
\sao{Additionally, results from this ablation study underline the significant roles played by the local and global branches. Using the basic BYOL setup without these branches as a baseline, we were able to demonstrate their critical contributions to enhancing the model's gaze estimation accuracy \cite{c17}.}

\newcolumntype{g}{>{\columncolor{gray}}c}
\begin{table}[ht]
\caption{Angular error ($\mathcal{L}_{ang}$) performance of various ablated SLYKLatent models: w/o-PMN denotes models without the patch module network, w/o-SSL denotes models trained without self-supervised learning, w/o-inv-EV denotes training without the Inverse Explained Variance modification, w/o-mBYOL denotes aimplemented using the BYOL framework. The final column reports the average percentage increase in $\mathcal{L}_{ang}$ relative to SLYKLatent over all datasets.}

\label{Tab: Ablation Results}
\begin{adjustbox}{width=\columnwidth,center}
\begin{tabular}{lcccccc}
\hline

\multirow{1}{*}{\textbf{Method}} & \multirow{1}{*}{\textbf{MP}}  & \multirow{1}{*}{\textbf{G360}}  & \multirow{1}{*}{\textbf{G360}}  & \multirow{1}{*}{\textbf{G360}} & \multirow{1}{*}{\textbf{ETH}} & \multirow{1}{*}{${\mathcal{L}_{ang}}\uparrow$}\\
                                 &                               & ($\pm \SI{180}{\degree}$)       &($\pm \SI{90}{\degree}$)         & ($\pm \SI{20}{\degree}$)                               &  & $(\%)$    \\        
\hline
w/o-PMN                         & $\SI{4.52}{\degree}$          & $\SI{11.21}{\degree}$         & $\SI{10.82}{\degree}$     & $\SI{8.92}{\degree}$            & $\SI{4.15}{\degree}$              & $7.81$  \\
w/o-SSL                         & $\SI{4.15}{\degree}$          & $\SI{10.90}{\degree}$         & $\SI{10.54}{\degree}$     & $\SI{8.94}{\degree}$            & $\SI{4.52}{\degree}$              &  $6.3$ \\
w/o-inv-EV                      & $\SI{4.23}{\degree}$          & $\SI{11.63}{\degree}$         & $\SI{11.30}{\degree}$     & $\SI{10.11}{\degree}$           & $\SI{4.24}{\degree}$              & $13.01$\\
w/o-mBYOL                       & $\SI{4.22}{\degree}$          & $\SI{10.82}{\degree}$         & $\SI{10.91}{\degree}$     & $\SI{9.61}{\degree}$            & $\SI{4.26}{\degree}$              & $8.41$ \\
\sao{w/o-local}                 & $\SI{5.53}{\degree}$          & $\SI{19.33}{\degree}$         & $\SI{18.87}{\degree}$     & $\SI{16.72}{\degree}$           & $\SI{7.22}{\degree}$              & $84.23$ \\
\sao{wo-global}                 & $\SI{3.99}{\degree}$          & $\SI{11.01}{\degree}$         & $\SI{10.67}{\degree}$     & $\SI{8.58}{\degree}$            & $\SI{4.11}{\degree}$              & $4.43$ \\
Basic BYOL                      & $\SI{4.56}{\degree}$          & $\SI{13.56}{\degree}$         & $\SI{12.50}{\degree}$     & $\SI{11.59}{\degree}$           & $\SI{5.00}{\degree}$              & $28.5$\\

\hline
{SLYKLatent}    & \textbf{\SI{3.77}{\degree}}                   & \textbf{\SI{10.70}{\degree}}  & \textbf{\SI{10.34}{\degree}}  & \textbf{\SI{8.03}{\degree}}           & \textbf{\SI{3.89}{\degree}} & -\\
\hline

\end{tabular}
\end{adjustbox}

\end{table}%

\subsubsection{Testing for Appearance Uncertainties}
\begin{figure}[ht]
    \centering
    \includegraphics[width=.47\textwidth]{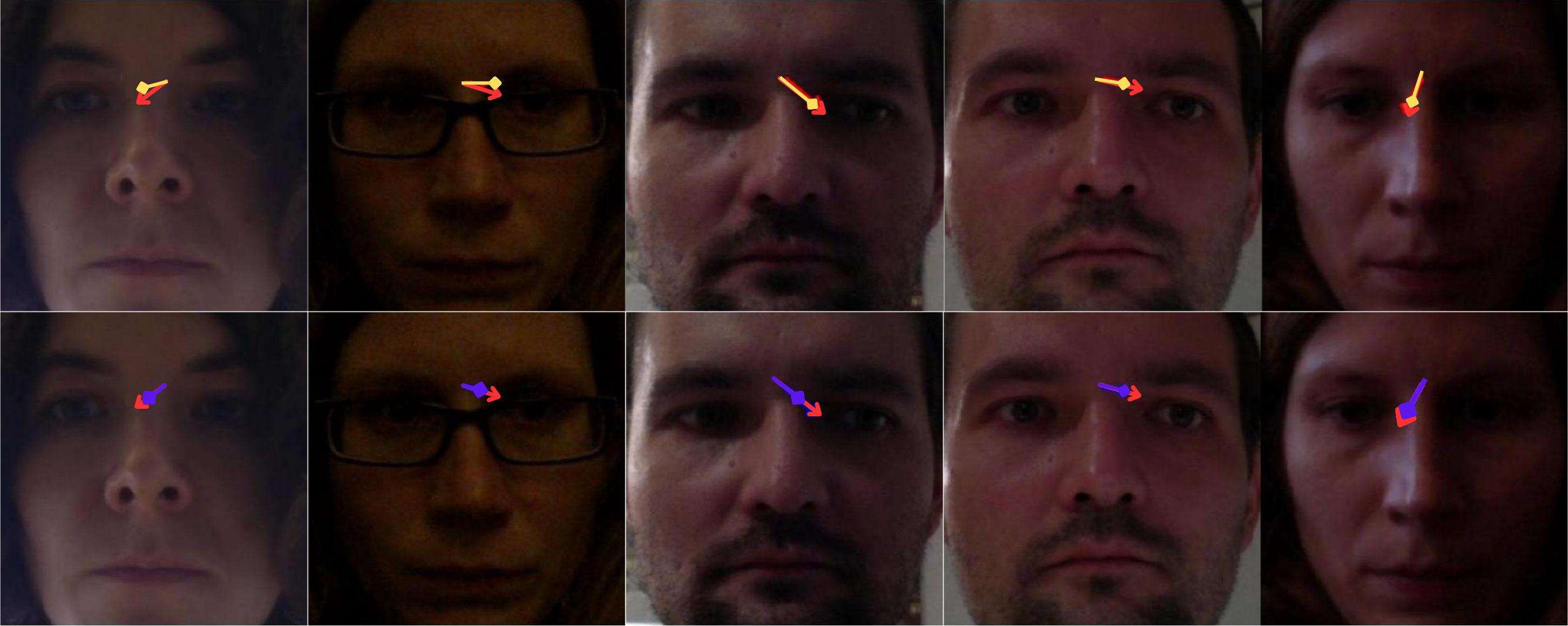}
    \caption{Comparison of gaze estimation under appearance uncertainties with L2CSNet. The top row shows predictions from the L2CSNet model, while the bottom row displays predictions from our SLYKLatent model. Various appearance uncertainties, such as low illumination and image blurriness, are demonstrated. \textcolor{red}{Red} arrows indicate ground truth gaze directions, \textcolor{yellow}{Yellow} arrows represent predictions from the L2CSNet model, and \textcolor{blue}{Blue} arrows represent predictions from the SLYKLatent model.}
    \label{fig:Gaze-plots-appearance}
\end{figure}

The robustness of gaze estimation models is paramount, particularly when facing real-world challenges characterized by appearance uncertainties, including, but not limited to, low illumination conditions. We conducted a rigorous examination of the MPIIFace dataset to pinpoint low-quality images indicative of such challenges. Our scan identified 98, 686, and 1223 low-illumination images within subsets p03, p11, and p13, respectively. 
We then subjected ablated variants of the SLYKLatent model (w/o-SSL, w/o-PMN, and w/o-inv-EV) to tests on these identified images to evaluate the individual contribution of each component of the SLYKLatent model to robustness to appearance uncertainties. The results of these tests are quantitatively summarized in Table \ref{Appearance Uncertainties Result}. \sao{Additionally, to complement these quantitative results, Figure \ref{fig:Gaze-plots-appearance} visually contrasts gaze estimation predictions under appearance uncertainties, mostly showcasing the performance of SLYKLatent against L2CSNet. The figure highlights the consistency of our model's predictions (blue arrows) under some challenging conditions compared to L2CSNet's (yellow arrows), with red arrows indicating the ground truth gaze directions.}

\begin{table}[htb]
\caption{Angular error performance of SLYKLatent (SLYKL) and it ablated variants for appearance uncertainties}
\label{Appearance Uncertainties Result}
\begin{adjustbox}{width=\columnwidth,center}
\begin{tabular}{lcccccc}
\hline
\textbf{Participants} & \textbf{No of Images}  &\textbf{w/o-PMN}  & \textbf{w/o-SSL}  & \textbf{w/o-inv-EV}  & \textbf {SLYKL} \\
\hline
P03      & $98$                       & $\SI{3.46}{\degree}$      & $\SI{2.96}{\degree}$            &$\SI{3.03}{\degree}$                 & $\SI{2.72}{\degree}$ \\
P11      & $1223$                     & $\SI{4.61}{\degree}$       &$\SI{4.26}{\degree}$           &$\SI{5.10}{\degree}$                  & $\SI{3.65}{\degree}$ \\ 
P13      & $686$                      & $\SI{4.53}{\degree}$       & $\SI{4.19}{\degree}$             & $\SI{7.53}{\degree}$               & $\SI{3.71}{\degree}$ \\
\hline
{$\mathcal{L}_{ang}\uparrow(\%)$} & - & $25.0$                 & $13.19$                          & $55.30$                                 & - \\
\hline
\end{tabular}
\end{adjustbox}
\end{table}
The improvements achieved by SLYKLatent when compared with the ablated models is substantial for all models ($25\%$, $13.19\%$ and $55.3\%$), and considerably greater than the improvements observed for the full MPIIFace dataset presented in Table \ref{Tab: Ablation Results} ($19.89\%$, $10.08\%$ and $12.20\%$). Most notably, for the selected high appearance uncertainty images the absence of inv-EV results in a $55.30\%$ increase in $\mathcal{L}_{ang}$ and the absence of PMN results in an increase of $25.0\%$. This suggests that both components are critical to the robustness of SLYKLatent and its ability to mitigate challenges due to appearance uncertainties such as low illumination. Similarly, the w/o-SSL configuration results in a $13.19\%$ increase in $\mathcal{L}_{ang}$. Although this is lower than the other enhancements, it shows that self-supervised learning offers a certain degree of resilience against environmental variances. 

\vspace{-7mm}
\subsection{Analysis of Equivariance Through Rotational Testing}
In this section, we investigate the equivariance properties of SLYKLatent, using the MPIIFace dataset for the analysis. The images, patches, and gaze vectors for this dataset are rotated to test for equivariance.

\textbf{Gaze Rotation:}
The groundtruth gaze vectors \(\vec{y}_{i}\) were rotated by an angle \(\theta\) using the equation:
\begin{equation}
    \vec{y}_{\text{i(rotated)}} = R(\theta) \vec{y}_{i},
\end{equation}
where \( R(\theta) \) is the rotation matrix defined as:
\begin{equation}
    R(\theta) = \begin{pmatrix}
        \cos(\theta) & -\sin(\theta) \\
        \sin(\theta) & \cos(\theta)
    \end{pmatrix}.
\end{equation}

\begin{figure}[ht]
    \centering
    \includegraphics[width=\linewidth]{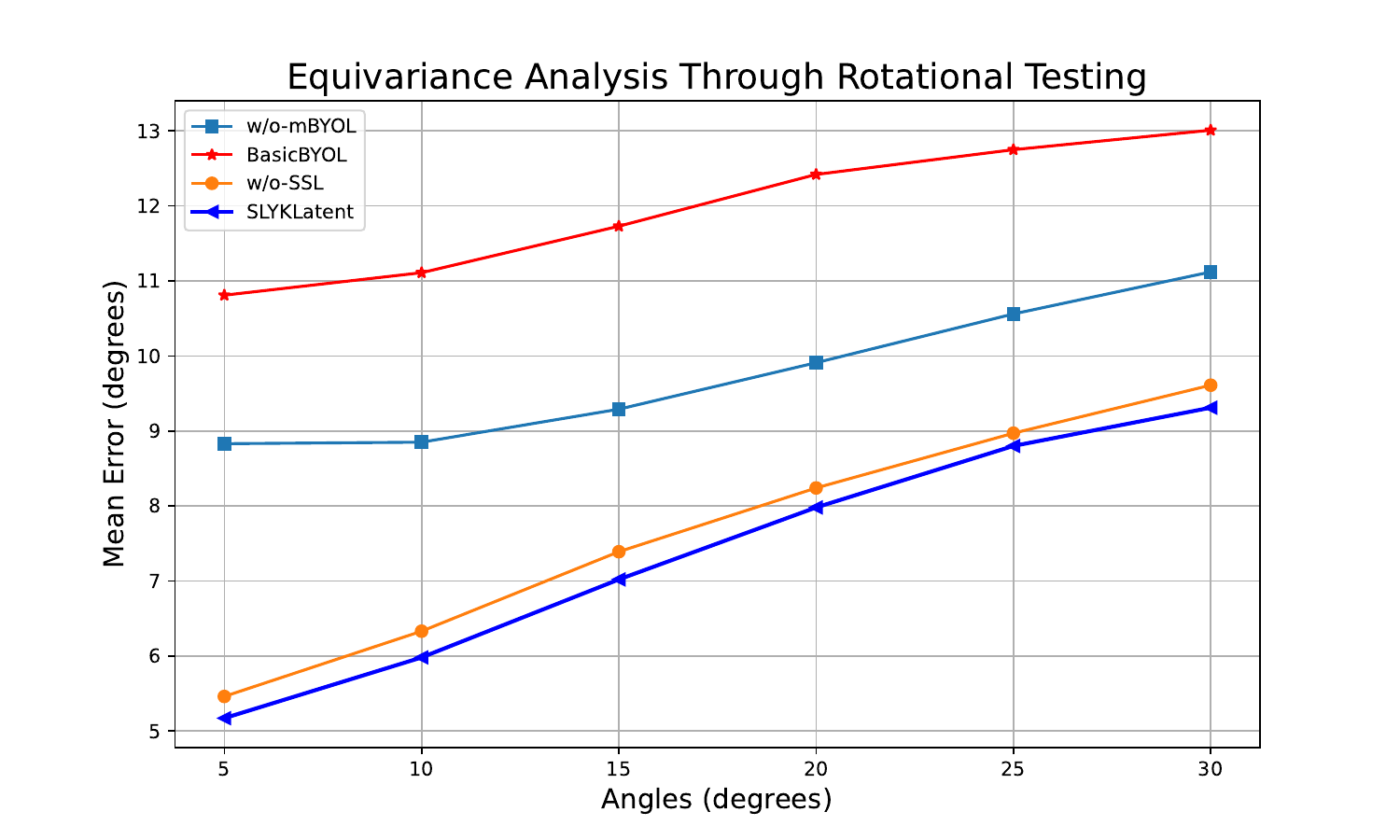}
    \caption{Equivariance performance of SLYKLatent and its ablated variants, w/o-mBYOL, w/o-SSL, and BasicBYOL, for rotational transformations}.
    \label{fig:equivariance_analysis}
\end{figure}%

\textbf{Results and Observations:}
SLYKLatent and selected ablated variants were subjected to this rotational transformation for rotation angles $\theta \in \{\SI{5}{\degree}, \SI{10}{\degree}, \SI{15}{\degree}, \SI{20}{\degree}, \SI{25}{\degree}, \SI{30}{\degree}\}$. The angular error results obtained are plotted in Fig. \ref{fig:equivariance_analysis} as a function of $\theta$. The Basic BYOL model has the highest angular errors followed by w/o-mBYOL, with errors in the range $\SI{10.8}{\degree}-\SI{13.0}{\degree}$ and  $\SI{8.8}{\degree}-\SI{11.2}{\degree}$, respectively.  In contrast, SLYKLatent is the best performing method with angular error rates ranging from $\SI{5.2}{\degree}$ to $\SI{9.3}{\degree}$. While w/o-SSL performs much better than the other ablated variants, it is consistently outperformed by SLYKLatent by between $\SI{0.2}{\degree}$ and $\SI{0.5}{\degree}$. To further elucidate SLYKLatent's equivariance properties, it is worth noting that the difference in angular error from $\SI{0}{\degree}$ (i.e. when not rotated) to $\SI{5}{\degree}$ is $\SI{6.3}{\degree}$ for BasicBYOL, $\SI{4.6}{\degree}$ for w/0-mBYOL, $\SI{1.4}{\degree}$ for w/o-SSL, and $\SI{1.2}{\degree}$ for SLYKLatent. These differences highlight the relative stability of SLYKLatent in the face of rotational transformations.

\textbf{Implications:}
The results make a compelling case for the importance of our mBYOL modifications, which include a modified transformation and a specialized branch network in the self-supervised pretraining stage. The substantial improvement in error rates in the SLYKLatent model, as compared to the Basic BYOL and without mBYOL configurations, validates our architectural decisions. Specifically, it confirms that our innovations contribute significantly to the model's resilience against rotational transformations, thereby affirming its robust equivariance capabilities. 

\subsection{Performance of SLYKLatent in Facial Expression Recognition (FER):}
Although the core objective of our study is gaze estimation, we extend the scope to assess SLYKLatent's adaptability in the domain of FER. This evaluation is relevant as FER, like gaze estimation, relies on the precise extraction of facial features. To this end, we repurpose our gaze estimation MLP to output class labels for facial expressions, specifically 7 classes for RAF-DB and 8 for AffectNet. Both datasets employ the weighted cross-entropy loss based on class weights.

On the RAF-DB and AffectNet benchmarks, SLYKLatent yields accuracy rates of 86.4\% and 60.9\%, respectively. While these figures fall slightly short of the performance of the state-of-the-art method cited in \cite{c64}, which employs a sophisticated Identity and Pose Disentangled Facial Expression Recognition (IPD-FER) model for disentangling facial expressions from identity and head pose, the results are still promising.  The IPD-FER model achieved an accuracy of 87.22\% for RAF-DB and 63.77\% for AffectNet, benefiting from its targeted disentanglement approach. However, it is worth noting that SLYKLatent demonstrates remarkable versatility, showing robustness not only in gaze estimation but also in FER, further showcasing its applicability to a wide range of vision tasks.

\section{Conclusion}
In this paper, we have presented SLYKLatent, an innovative framework designed to tackle the intricate challenges inherent to gaze estimation, including appearance instability, covariate shift, and the need for domain generalization. SLYKLatent integrates the resilience of Self-Supervised Learning (SSL) with a specialized tri-branch, patch-centric network architecture (PMN) and enhanced training via an inverse explained variance (inv-EV) weighted loss function. Several ablation studies presented provide compelling empirical insights into the contribution of these different components to the overall performance of SLYKLatent.

Through extensive evaluation on diverse datasets (MPIIFaceGaze, Gaze360, and ETHX-Gaze) and benchmarking against several competing methods in the literature, we demonstrate that SLYKLatent achieves state-of-the-art performance for gaze estimation. In addition, we show that the resulting model manifests a notable flexibility in addressing tasks beyond gaze estimation, such as facial expression recognition. Further tests on MPIIFace show that the PMN and inv-EV components of SLYKLatent contribute to its robust to rotations although a comprehensive benchmark for this specific capability is an avenue for future work.

SLYKLatent does have some limitations. In particular, its effectiveness is contingent upon reliable eye patch detection, hence it cannot be employed in scenarios where such features are not consistently present, as was the case for $65\%$ of the ETHZ-Gaze dataset. Additionally, while the model demonstrates promising versatility in tasks like facial expression recognition, it has yet to meet the gold standards established by domain-specific state-of-the-art models.

In future work, we aim to broaden the scope of SLYKLatent's applicability by exploring other facial feature estimation tasks and by examining its integrative potential in complex systems, such as human-robot interaction platforms. A further potential refinement is the addition of Bayesian Neural Networks to extend the model's capabilities to further capture aleatoric uncertainties in the learned features. These enhancements will contribute to a more nuanced understanding of the framework's capabilities and limitations, thereby facilitating its more effective deployment in real-world scenarios.

\ifCLASSOPTIONcompsoc
  \section*{Acknowledgments}
\else

\section*{Acknowledgment}
\fi
This research is funded by the Engineering and Physical Sciences and Research Council (EPSRC), United Kingdom. We are grateful for the use of the computing resources from the Northern Ireland High Performance Computing (NI-HPC) service funded by EPSRC (EP/T022175).
%
%
%
\ifCLASSOPTIONcaptionsoff
  \newpage
\fi

\vspace{-10mm}
\begin{IEEEbiography}[{\includegraphics[width=1in,height=1.8in,clip,keepaspectratio]{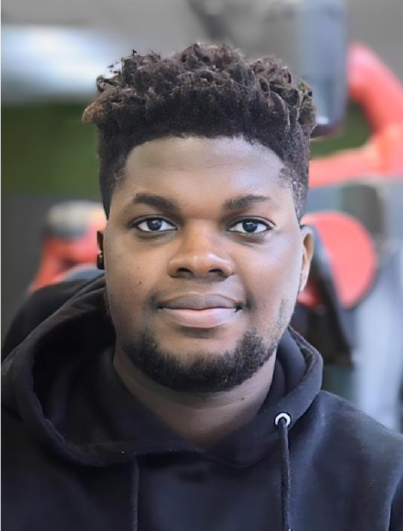}}]{Samuel Adebayo}
earned a First Class BSc in Electronic and Computer Engineering from Lagos State University, Lagos Nigeria (2016-2020) and is currently pursuing his PhD in Machine Learning at Queen's University Belfast, UK (2020-2024). Samuel's research is deeply rooted in computational intelligence, with a primary focus on the intricate dynamics of human hand-eye coordination and its implications for human-robot interactions. This aligns with his broader interest in advancing the fields of Computer Vision, Deep Learning, Natural Language Processing, and Bayesian Machine Learning. Beyond academics, Samuel transitioned from his role as the branch chair of the IEEE Student Branch at Lagos State University to his current leadership position as the Student Branch Chair at Queen's University Belfast.
\end{IEEEbiography}
\vspace{-10mm}
\begin{IEEEbiography}[{\includegraphics[width=1in,height=1.8in,clip,keepaspectratio]{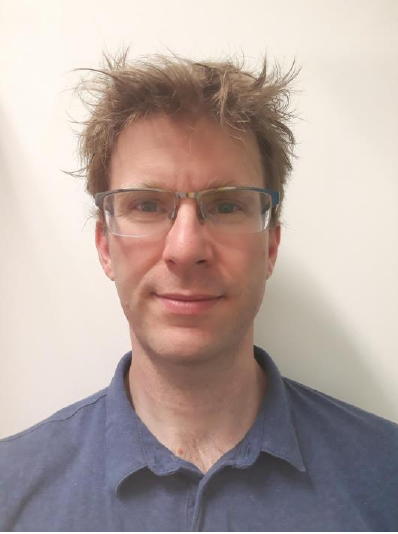}}]
{Joost C. Dessing} received MSc. And PhD degrees in Human Movement Sciences from Vrije University Amsterdam, Amsterdam, The Netherlands, in 2001 and 2005, respectively.
He is currently a Lecturer at the School of Psychology of Queen's University Belfast, where he co-directs the Science in Motion lab. His research interests are fundamental and applied instances of eye-hand coordination, with a particular focus on sports and manufacturing scenarios. He studies these using 3D motion tracking, virtual reality, and neurophysiological tools.
\end{IEEEbiography}
\vspace{-10mm}
%
\begin{IEEEbiography}[{\includegraphics[width=1in,height=1.8in,clip,keepaspectratio]{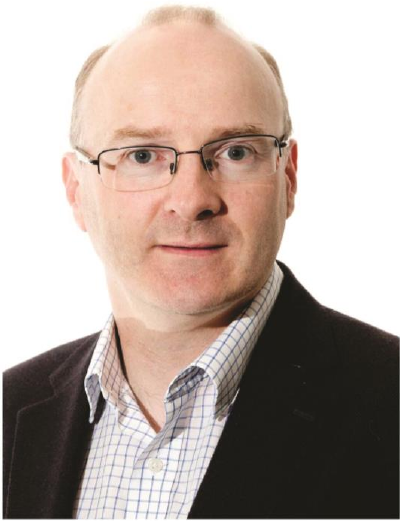}}]
{Se\'an McLoone} (S$'94$ -- M$'88$  -- SM$'02$) received an M.E. degree in Electrical and Electronic Engineering and a PhD in Control Engineering from Queen's University Belfast, Belfast, U.K. in 1992 and 1996, respectively.
He is currently a Professor and Director of the Energy Power and Intelligent Control Research Centre at Queen's University Belfast. His research interests are in Applied Computational Intelligence and Machine Learning with a particular focus on data based modelling and analysis of dynamical systems, with applications in advanced manufacturing informatics, energy and sustainability, connected health and assisted living technologies.
Prof. McLoone is a Chartered Engineer and Fellow of the Institution of Engineering and Technology. He is a Past Chairman of the UK and Republic of Ireland (UKRI) Section of the IEEE.
\end{IEEEbiography}
\clearpage
\onecolumn
\section*{Supplementary Material}
\begin{spacing}{1.2} 




\section*{Introduction}
This supplementary material provides detailed descriptions of the datasets used in our study, \sao{"SLYKLatent: A Learning Framework for Gaze Estimation Using Deep Facial Feature Learning".} These descriptions are intended to offer comprehensive insights into the datasets' specifics, aiding in the understanding of their application and significance in our research. Additionally, we have also present the visual depiction of gaze plots of SLYKLatent against mainstream methods.

\section{Detailed Dataset Descriptions}
For SSL pre-training we used the AFFECTNet Dataset \cite{c52}. These image samples are already aligned, however, we cropped out eye patches which are then used as input for the corresponding local branches. For downstream evaluation, MPIIFaceGaze\cite{c34}, Gaze360 \cite{c39}, and ETHX-Gaze \cite{c53} were used for the evaluation of the performance of gaze estimation. Furthermore, to demonstrate the robustness of our approach, we also evaluated the performance of SLYKLatent on facial expression recognition datasets including the AFFECTNet \cite{c52} and RAF-DB datasets \cite{c51}. During inference and downstream fine-tuning training, each data sample consists of the eye patches (left and right) and the full face images.

\textbf{Gaze360} \cite{c39} is a gaze estimation dataset that captures the gaze of subjects in a naturalistic setting. The dataset includes images of participants' faces in a variety of poses and lighting conditions. The gaze data in Gaze360 was collected using a high-precision Pupil Labs camera. The dataset contains over 172K images with the corresponding gaze vectors in 2D and 3D, it also includes facial landmarks, head pose, and 3D target location. The dataset has a training, validation, and test set. The data distribution is not balanced and there is a bias towards certain gaze directions. In the experiments we trained the model on data samples within the full range ($\pm \SI{180}{\degree}$) and tested it on both the front \SI{180}{\degree} (i.e. $\pm \SI{90}{\degree}$) and the front $\SI{40}{\degree}$ ($\pm \SI{20}{\degree}$) samples.

\textbf{MPIIFaceGaze} \cite{c34} is an extension of the MPIIGaze dataset \cite{c33}, which encompasses a wide range of data, including eye patches, gaze targets, facial landmarks, annotations indicating the center of the eye pupil and comprehensive facial images. This dataset was generated through the contributions of 15 participants and contains a total of 45,000 samples. Each sample within the dataset consists of a facial image, with a resolution of $448 \times 448$ pixels, accompanied by a suite of annotations. These annotations include 2D and 3D gaze angle vectors, head pose in 3D space, the centre of the face, the 3D location of the gaze target, and six distinguishable facial landmarks. In view of the considerable computational load associated with processing the high-resolution images on the GPU during the training phase, we found it pragmatic to resize all images to a more digestible resolution of $224 \times 224$ pixels. In terms of evaluation, we opted for the leave-one-person-out-cross-validation method, as delineated in the original paper\cite{c34}. This method entails the systematic exclusion of one participant's data during each training cycle, with the model being trained on the remaining pool of 42,000 samples (comprising 14 participants' data). At every training cycle, we used the last 3000 samples of the 42000 for validation. 

\textbf {ETH-XGaze} \cite{c53} provides more than 1M+ data points that capture a broad range of positions and gaze angles. With the aim of overcoming the existing limitations associated with the dearth of large-scale gaze datasets, ETHX-gaze was collected from a diverse group of 110 participants. The dataset includes high-resolution face images in two sizes ($224 \times 224$ and $448 \times 448$ pixels) and corresponding gaze vectors. For our experiment, we utilized the $224 \times 224$ pixel version of the dataset. As the dataset lacks facial landmarks, which are necessary for our eye patch methodology, we employed dlib's facial landmark detection algorithms \cite{c57}. This led to the exclusion of images without detectable eye patches, reducing our image pool to approximately 360k. From this subset, we designated 10\% for validation and 10\% for testing.

\textbf{RAF-DB} \cite{c51} is a dataset of real-world facial expressions created for the task of facial expression recognition. It features a collection of 29,672 real-world images showcasing various facial expressions. These images are labeled with categorical expressions such as angry, happy, neutral, sad, disgust, surprise, and fear. The dataset encompasses a broad spectrum of poses, illuminations, and occlusions, providing a rich resource for training and testing facial expression recognition models.

\textbf{AFFECTNet} \cite{c52}, a dataset created for facial expression recognition (FER) and affect analysis, contains more than 270K images of faces with various facial expressions. Faces in the dataset are labelled with categorical expressions (angry, happy, neutral, sad, disgust, surprise, and fear) and valence-arousal continuous values. AFFECTNet stands out due to its high degree of diversity in terms of participant age, gender, and ethnicity. It also contains a large proportion of non-frontal and occluded face images, providing a comprehensive dataset for robust FER model training. However, a degree of imbalance exists within the dataset. To account for this, we computed the weighted average of each class with respect to the total sample size.

In summary, the datasets employed encompass a broad spectrum of gaze estimation and facial expression recognition tasks, incorporating images gathered in controlled lab settings as well as real-world environments. They also represent participants from a diverse range of ages, genders, and ethnicities. This comprehensive variety is crucial for robustly assessing the effectiveness of our methodology.

\section{Gaze Plots Comparison}
\sao{We replicated the model architecture of L2CS-Net \cite{c15} and MANet \cite{c59} and compared their plots to ours.

\begin{figure}[h]
    \centering
    \includegraphics[width=\textwidth]{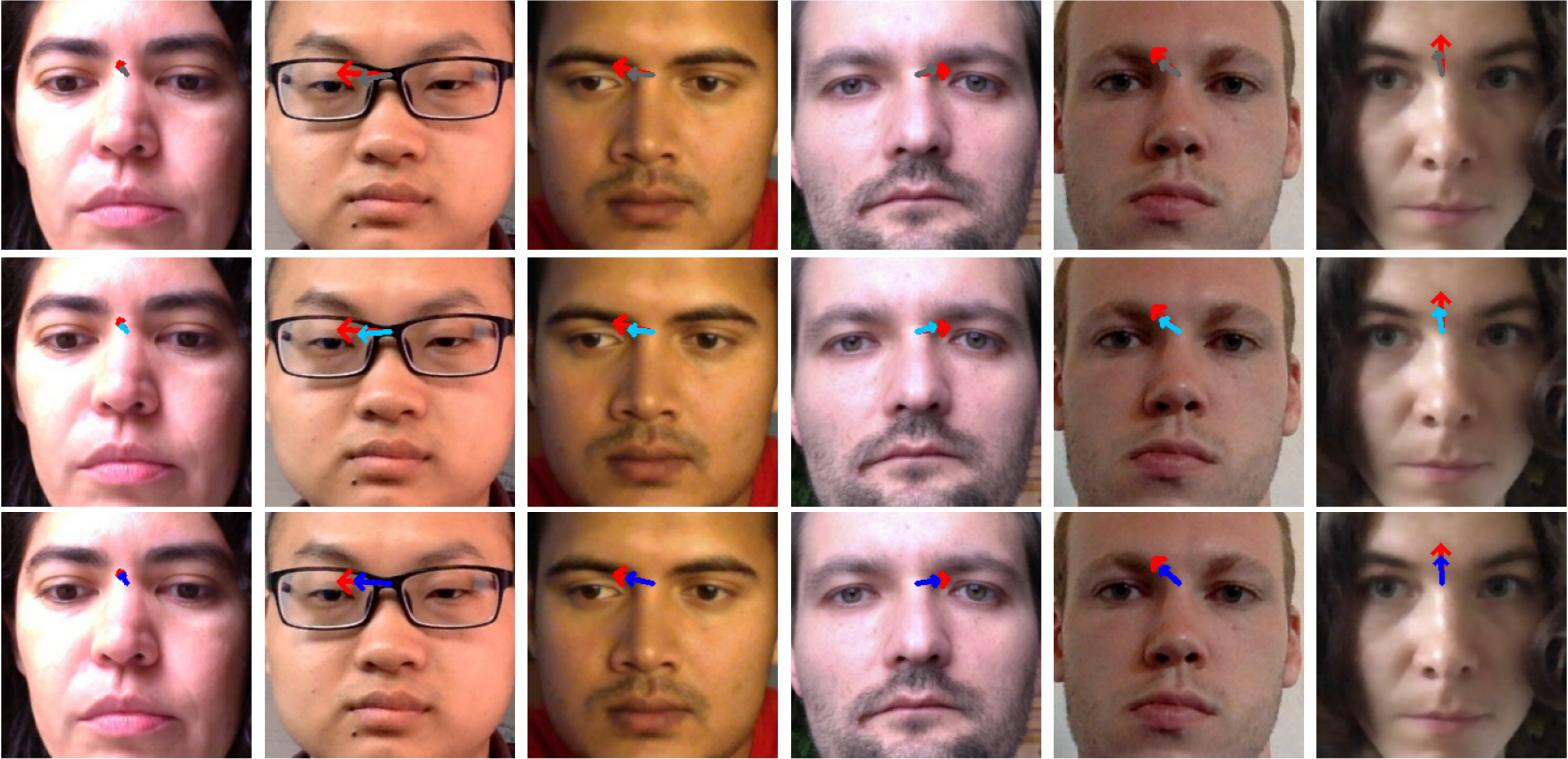} 
    \caption{Comparison of gaze direction estimations across MaNet (top row), L2CS-Net (middle row), and SLYKLatent (bottom row) models. SLYKLatent demonstrates higher stability and precision in gaze estimation across varied facial inputs.}
    \label{fig:gaze_comparison}
\end{figure}

Figure \ref{fig:gaze_comparison} illustrates the predicted gaze directions for each model across multiple facial images, organized by row as follows:
\begin{itemize}
    \item \textbf{Top row} - predictions by MaNet.
    \item \textbf{Middle row} - predictions by L2CS-Net.
    \item \textbf{Bottom row}  - predictions by SLYKLatent.
\end{itemize}
In the visualization, each arrow represents the estimated gaze direction, with different colours used to aid comparisons. The arrow in \textcolor{red}{red}  in each visualization denotes the groundtruth gaze direction. The accuracy of each model's predictions can be assessed by the proximity of its estimated gaze direction to the ground truth across various facial images. SLYKLatent consistently aligns more closely with the ground truth, indicating higher precision in its estimation performance across different facial inputs.

While MaNet and L2CS-Net provide reasonable estimates, slight deviations are visible in some directions, likely due to variations in head pose and lighting conditions. These inconsistencies lead to less reliable predictions in more challenging input scenarios. In contrast, our SLYKLatent model demonstrates higher stability and accuracy, consistently providing robust predictions across varied face orientations. This stability can be attributed to SLYKLatent’s architecture, which is optimized to capture subtle gaze cues, leading to reduced drift and increased alignment with expected gaze directions.}

\end{spacing}
\end{document}